\documentclass[]{bytedance_seed}

\usepackage[toc,page,header]{appendix}

\usepackage{minitoc}
\usepackage{amsmath}
\usepackage{amssymb}
\usepackage{mathtools}
\usepackage{colortbl}
\usepackage{url}
\usepackage{float}

\newcommand{\parhead}[1]{\vspace{0.35em}\noindent\textbf{#1.}\ }

\title{LoGo: Token-Level Dynamic Local-Global Attention}

\author[1,2]{Yuqi Pan}
\author[2,\dag,*]{Zheng Li}
\author[2]{Bohao Tang}
\author[2]{Zhen Qin}
\author[1,*]{Guoqi Li}
\affiliation[1]{Institute of Automation, Chinese Academy of Sciences}
\affiliation[2]{ByteDance Seed}
\contribution[\dag]{Project lead}
\contribution[*]{Corresponding authors}

\abstract{As context lengths scale, attention increasingly becomes a primary computational bottleneck in large language models. Standard Transformers remain powerful but computationally inefficient, as they allocate the same attention budget to every token regardless of its contextual demand. Existing local-global hybrids provide a more efficient alternative by mixing restricted- and full-context attention, but they typically allocate span statically across layers or heads. To address these limitations, we propose LoGo, a token-level dynamic local-global attention mechanism that uses attention span as a direct proxy for attention budget allocation. Each LoGo layer contains coupled local and global branches: all tokens receive efficient local attention over a restricted context window, while a learned gate activates global attention with full-context access only for tokens requiring long-range information. A threshold-based budget controller maintains a target global ratio without auxiliary losses, and a progressive masking schedule stabilizes training before sparse routing takes effect. We further implement query-sparse Triton kernels that convert reduced global-attention computation into practical speedups. Extensive experiments validate LoGo's effectiveness, showing that it preserves the scaling behavior of full-attention Transformers across model sizes. In controlled comparisons, LoGo improves over the full-attention Transformer and matched-budget static local-global hybrids, with clear gains on long-range retrieval. Analysis further shows that LoGo learns interpretable span allocation patterns. These results suggest that learned token-level span allocation is an effective and scalable way to improve the long-context performance-compute trade-off.}

\date{\today}
\correspondence{\email{panyuqi2024@ia.ac.cn}, \email{lizheng.m@bytedance.com}, \email{guoqi.li@ia.ac.cn}}

\begin{document}
\maketitle


\section{Introduction}

Large language models (LLMs) have advanced rapidly through continual improvements in performance under fixed compute budgets. As context lengths scale, attention becomes an increasingly dominant source of computational cost. Yet standard Transformers~\citep{vaswani2017attention} allocate this computation uniformly across tokens, despite substantial variation in prediction difficulty and contextual demand. Conditional computation~\citep{bengio2013estimating} offers a natural way to address this mismatch by allocating compute on demand. In long-context modeling, this motivates dynamic attention-budget allocation across tokens as a path toward more efficient long-context scaling.

The key challenge is how to estimate the attention budget to assign to each token before attention is computed. We argue that attention span is a natural proxy for this budget: it controls both accessible context and cost, and its demand can often be anticipated from linguistic context. Tokens in fixed phrases may be predictable from local evidence, whereas tokens involved in long-range retrieval require global access. However, existing local-global hybrids allocate span statically, either across layers (inter-layer hybrids; \citep{team2024gemma,agarwal2025gpt}) or across heads (intra-layer hybrids; \citep{donghymba}). In both cases, all tokens follow the same span pathway regardless of their contextual needs. Such static layouts impose preset per-token budgets, often wasting global computation on locally predictable tokens. Existing dynamic variants remain limited, as they typically switch among attention mechanisms rather than allocate span directly~\citep{ren2023sparse}, rely on handcrafted routing~\citep{li2026transmamba,li2026multi}, or lack matched-budget evaluation at scale~\citep{ainslie2023colt5}. These gaps motivate three questions:
\begin{itemize}
    \item[\textbf{RQ1.}] Can a model learn token-level local/global span allocation, spending global attention on tokens that require long-range context?
    \item[\textbf{RQ2.}] As a pretrained backbone, can such a model preserve the scaling behavior of standard Transformers?
    \item[\textbf{RQ3.}] Under matched parameters and FLOPs, can token-level span allocation improve the performance-compute trade-off over static local-global hybrids?
\end{itemize}

\begin{figure}[!t]
\centering
\vspace{-1cm}
\includegraphics[width=0.8\textwidth]{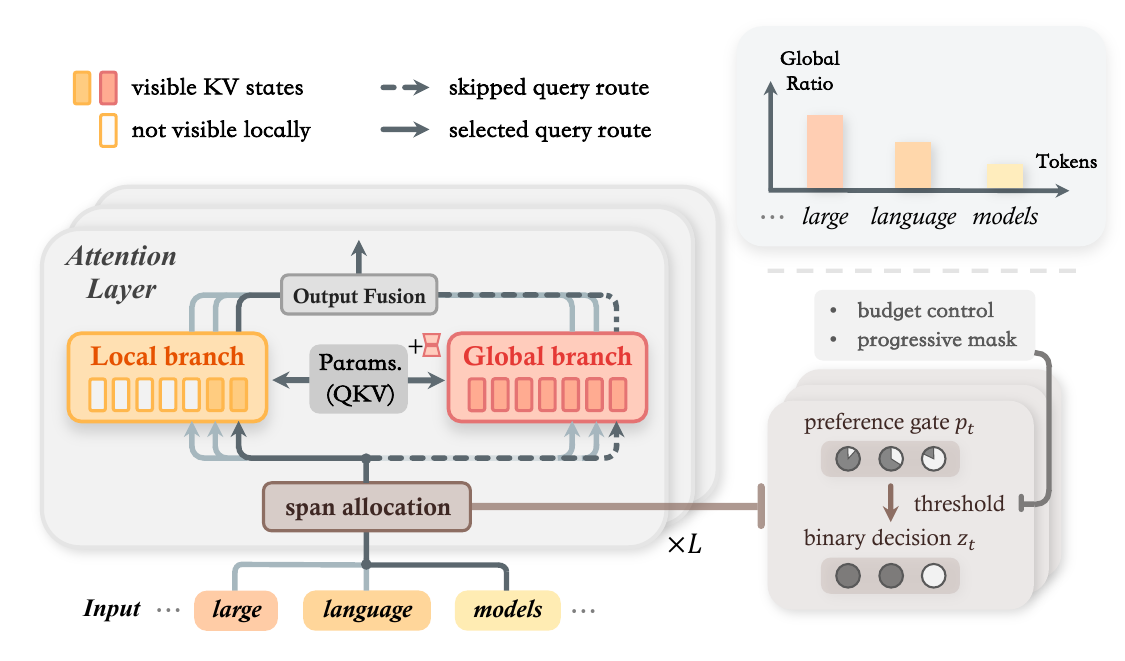}
\caption{\textbf{Overview of LoGo.} Each layer applies local attention to all tokens while activating the global branch only for selected query routes. A token-level gate predicts global-span preference, an adaptive threshold converts this preference into binary routing decisions under a target global ratio, and P-mask gradually introduces sparse routing during training.}\label{fig:logo-overview}
\end{figure}

To answer these questions, we propose LoGo (Looking far, Glancing only), a token-level dynamic local-global attention mechanism. Figure~\ref{fig:logo-overview} illustrates the overall design. Each LoGo layer contains local and global branches with the same attention form but different accessible spans, sharing the main attention parameters while being decoupled by lightweight transformations. A learned scalar gate predicts each token's preference for global span: selected tokens combine local and global outputs, whereas unselected tokens bypass the global branch and remain purely local. The gate is compared with an adaptive threshold that maintains a target global activation ratio, and a progressive masking schedule delays sparse routing until the gate has learned meaningful preferences. Because LoGo sparsifies only global queries while preserving all key-value states, selected tokens retain exact full-context access. We further implement query-sparse Triton kernels that translate reduced global-attention computation into practical speedups.

Extensive experiments validate LoGo's effectiveness. First, scaling runs from 200M to 3.3B parameters show that LoGo consistently matches or improves standard Transformers on language modeling, indicating that token-level span allocation preserves backbone scalability. Second, in controlled 1.5B comparisons under matched parameters and attention FLOPs, LoGo improves the performance-compute trade-off over the standard Transformer and static inter- and intra-layer hybrids: it achieves the best language-modeling results, remains competitive on short-context recall, and provides clear gains on long-range retrieval after context extension. Third, kernel benchmarks show that our implementation converts reduced FLOPs into practical speedups, with near-proportional gains at long sequence lengths. Additional ablations validate the main design choices, show robustness across global ratios, and demonstrate compatibility with static hybrids for reducing both memory and computation. Finally, analysis shows that LoGo learns interpretable allocation patterns, assigning more global span to tokens requiring long-range access.

\section{Related Work}

\parhead{Conditional Computation in Transformers}
Conditional computation improves the performance-compute trade-off through token-adaptive compute budgets~\citep{bengio2013estimating,bengio2015conditional}. Existing methods dynamically allocate compute along several axes: experts~\citep{huang2024harder,yang2024xmoe,jinmoepp}, depth~\citep{elbayaddepth,raposo2024mixture}, and attention via adaptive head selection~\citep{zhang2022mixture} or token-wise sparse attention budgets~\citep{lin2026twilight}. The closest line dynamically activates attention modules. CoLT5~\citep{ainslie2023colt5} routes tokens through light and heavy branches for feed-forward and attention computation, but focuses on encoder-side routing. SMA~\citep{ren2023sparse} activates a GAU module based on SSM states, but activated tokens still share the same restricted span. Moreover, both are evaluated at limited scale without matched-budget comparisons. LoGo instead targets decoder-only backbones and learns token-level span allocation as a direct proxy for dynamic attention budgeting. We evaluate it at pretrained scale under matched parameters and FLOPs, with emphasis on long-context capability. A broader discussion is provided in Appendix~\ref{app:related-conditional-computation}.

\parhead{Local-Global Hybrid Attention}
Local-global hybrid attention reduces long-context cost by combining restricted-span computation with full-context access. Existing designs typically apply the mixture statically across layers, as in inter-layer hybrids~\citep{team2024gemma,agarwal2025gpt,xiao2026mimo}, or across heads, as in intra-layer hybrids~\citep{donghymba,zuo2025falcon}. Some sequence-adaptive hybrids vary attention pathways across positions, but most use fixed predefined patterns~\citep{mcdermott2025lola}, while closer methods such as TransMamba~\citep{li2026transmamba} and Oryx~\citep{li2026multi} rely on handcrafted per-token budgets rather than learned routing. In contrast, LoGo keeps the attention form fixed and learns span allocation end-to-end: each token decides whether to activate full-context attention under a controlled global budget. This enables matched-budget comparison with static hybrids while avoiding unnecessary global computation on locally predictable tokens. A broader discussion is provided in Appendix~\ref{app:related-local-global-hybrids}.

\section{Methods}
\label{sec:methods}

We present LoGo in three parts. Section~\ref{sec:preliminaries} formalizes global/local attention and the token-level span-allocation objective. Section~\ref{sec:logo-attention} introduces the LoGo attention layer and its three core components. Section~\ref{sec:efficient-implementation} describes the query-sparse implementation for reduced computation.

\subsection{Preliminaries and Problem Formulation}
\label{sec:preliminaries}

\parhead{Global and Local Attention}
Let $\mathbf{X}^{(\ell)}=[\mathbf{x}^{(\ell)}_1;\ldots;\mathbf{x}^{(\ell)}_T]\in\mathbb{R}^{T\times d}$ denote the input hidden states to the $\ell$-th attention layer, and omit $\ell$ when clear. Standard multi-head attention projects $\mathbf{X}$ into $\mathbf{Q},\mathbf{K},\mathbf{V}$ and computes the attention output for each head as:
\begin{equation}
\mathbf{o}_t
=\sum_{s\in\mathcal{C}_t} a_{t,s}\mathbf{v}_s
=\sum_{s\in\mathcal{C}_t}
\frac{\exp(\mathbf{q}_t\mathbf{k}_s^\top/\sqrt{d_h})}
{\sum_{j\in\mathcal{C}_t}\exp(\mathbf{q}_t\mathbf{k}_j^\top/\sqrt{d_h})}
\mathbf{v}_s ,
\label{eq:attention}
\end{equation}
where $\mathcal{C}_t$ denotes the causal context accessible to query position $t$, and $d_h$ is the head dimension. Global attention~\citep{vaswani2017attention} uses the full causal prefix $\mathcal{C}^{\infty}_t=\{1,\ldots,t\}$, while local attention~\citep{beltagy2020longformer} restricts context to a fixed window $\mathcal{C}^{w}_t=\{\max(1,t-w+1),\ldots,t\}$. We denote the corresponding operators by $\operatorname{SoftmaxAttn}_{\infty}(\mathbf{Q},\mathbf{K},\mathbf{V})$ and $\operatorname{SoftmaxAttn}_{w}(\mathbf{Q},\mathbf{K},\mathbf{V})$, respectively. Global attention provides full-context access at $\mathcal{O}(T^2d)$ cost, whereas local attention reduces the cost to $\mathcal{O}(Twd)$ at the price of losing direct access to tokens outside the window.

\parhead{Static Local-Global Hybrids}
Let $z_{t,h}^{(\ell)}\in\{0,1\}$ indicate whether token $t$ uses global attention at layer $\ell$ and head $h$. Static hybrids fix this span decision independently of the token position. Inter-layer hybrids choose $z_{t,h}^{(\ell)}$ by layer, whereas intra-layer hybrids choose it by head. In both cases, $z_{t,h}^{(\ell)} = z_{h}^{(\ell)}$ for all $t$. Thus every token follows the same predefined local-global pathway, despite differences in contextual demand.

\parhead{Token-Level Span Allocation Objective}
LoGo learns token-level span allocation jointly with the language-modeling objective, aiming to minimize prediction loss under a prescribed attention-compute budget. At each layer, it learns to predict a binary routing decision:
\begin{equation}
z_t^{(\ell)}=f_{\theta}^{(\ell)}(\mathbf{x}_t^{(\ell)}),
\quad
z_t^{(\ell)}\in\{0,1\},
\label{eq:token-routing}
\end{equation}
where $z_t^{(\ell)}=1$ activates global attention for token $t$, and $z_t^{(\ell)}=0$ uses local attention only. All heads in a layer share this token-level decision for efficient implementation. To constrain the attention budget, we define a target global activation ratio $\rho^{(\ell)}$. The realized ratio at layer $\ell$ is:
\begin{equation}
F^{(\ell)}=\frac{1}{|\mathcal{B}|}\sum_{(b,t)\in\mathcal{B}} z_{b,t}^{(\ell)},
\label{eq:realized-ratio}
\end{equation}
where $\mathcal{B}$ denotes valid token positions in the training batch. LoGo enforces the budget by maintaining $F^{(\ell)}\approx\rho^{(\ell)}$, formulating dynamic attention-budget allocation as a constrained token-level choice between local and global span.

\subsection{LoGo Attention}
\label{sec:logo-attention}

LoGo implements token-level span allocation via three components: coupled local/global attention branches, a token-level global-preference gate, and a budget controller with progressive masking.

\parhead{Coupled Local and Global Branches}
Each LoGo layer contains two branches with the same attention form: a local branch over a short causal window ($w=128$ in our main experiments) and a global branch over the full causal context. To control parameter size while allowing branch specialization, we first use shared main attention projections to produce the local-branch representations:
\begin{align}
\mathbf{Q}^{L} = \mathbf{X}\mathbf{W}_q, \quad
\mathbf{K}^{L} = \mathbf{X}\mathbf{W}_k, \quad
\mathbf{V}^{L} = \mathbf{X}\mathbf{W}_v .
\end{align}
For each head $h$, the global representations are then obtained by applying a lightweight transformation to the corresponding local-branch representations, with $\mathbf{U}^{G}_h=\phi_{U,h}(\mathbf{U}^{L}_h)$ for $\mathbf{U}\in\{\mathbf{Q},\mathbf{K},\mathbf{V}\}$. We instantiate $\phi_{U,h}$ with either a linear projection (LP) or a scale-and-offset transformation (SO):
\begin{equation}
\begin{alignedat}{2}
\text{LP:}\quad
\phi_{U,h}(\mathbf{U}^{L}_h)
&=\mathbf{U}^{L}_h\mathbf{A}_{U,h},
&\qquad
\mathbf{A}_{U,h}&\in\mathbb{R}^{d_h\times d_h},\\
\text{SO:}\quad
\phi_{U,h}(\mathbf{U}^{L}_h)
&=\mathbf{U}^{L}_h\odot\mathbf{s}_{U,h}+\mathbf{b}_{U,h},
&\qquad
\mathbf{s}_{U,h},\mathbf{b}_{U,h}&\in\mathbb{R}^{d_h}.
\end{alignedat}
\label{eq:branch-transform}
\end{equation}
We use LP as the default in LoGo and include SO as an ablation. Both forms are lightweight relative to the main $\mathcal{O}(d^2)$ projections and add roughly 1\% parameters in our 1.5B setting. The transformations are initialized as identities, so the branches start from shared representations and specialize during training. The output projection $\mathbf{W}_o$ is shared.

The two branches then compute span-specific attention:
\begin{align}
\mathbf{O}^{L}
&=\operatorname{SoftmaxAttn}_{w}
(\mathbf{Q}^{L},\mathbf{K}^{L},\mathbf{V}^{L}),\quad
\mathbf{O}^{G}=\operatorname{SoftmaxAttn}_{\infty}
(\mathbf{Q}^{G},\mathbf{K}^{G},\mathbf{V}^{G}),
\label{eq:branch-attention}
\end{align}
where the global branch will be computed only for selected query positions, as described next.

\parhead{Token-Level Span Allocation and Output Fusion}
For each token, LoGo predicts a scalar global-context preference $p_t^{(\ell)}$ using a gate vector $\mathbf{w}_g^{(\ell)}$. A layer-specific threshold $b_{\mathrm{th}}^{(\ell)}$ converts this continuous preference into a hard global-activation decision $z_t^{(\ell)}$:
\begin{equation}
p_t^{(\ell)}
=\operatorname{sigmoid}\!\left({\mathbf{x}_t^{(\ell)}}^\top\mathbf{w}_g^{(\ell)}\right),
\quad
z_t^{(\ell)}=\mathbb{I}\!\left[p_t^{(\ell)}>b_{\mathrm{th}}^{(\ell)}\right].
\label{eq:gate-routing}
\end{equation}
With near-zero initialization of the gate projection, $p_t^{(\ell)}$ is initially centered around 0.5. Let $\mathcal{I}^{(\ell)}=\{t:z_t^{(\ell)}=1\}$ denote the selected global-query positions. The local branch covers all tokens, while the global branch is computed only for queries in $\mathcal{I}^{(\ell)}$:
\begin{equation}
\mathbf{O}^{G}_{\mathcal{I}^{(\ell)}}=
\operatorname{SoftmaxAttn}_{\infty}
\left(
\mathbf{Q}^{G}_{\mathcal{I}^{(\ell)}},
\mathbf{K}^{G},
\mathbf{V}^{G}
\right).
\label{eq:query-sparse-global}
\end{equation}
LoGo sparsifies only global queries while keeping $\mathbf{K}^{G}$ and $\mathbf{V}^{G}$ dense over all valid tokens, so selected queries retain exact full-context access.

The continuous gate also serves as the fusion weight between local and global outputs. Before fusion, branch-specific ContextNorm operators, implemented as head-wise RMSNorm following \citet{sun2023retentive}, normalize the output scale of each branch:
\begin{equation}
\widetilde{\mathbf{O}}^{L}=\operatorname{Norm}^{L}(\mathbf{O}^{L}),
\quad
\widetilde{\mathbf{O}}^{G}=\operatorname{Norm}^{G}(\mathbf{O}^{G}).
\label{eq:contextnorm}
\end{equation}
This prevents the gate from compensating for branch-wise magnitude mismatch and keeps it focused on learning span preference. The fused attention output is:
\begin{equation}
\mathbf{o}_t=
\begin{cases}
(1-p_t^{(\ell)})\cdot\widetilde{\mathbf{o}}^{L}_t
+p_t^{(\ell)}\cdot\widetilde{\mathbf{o}}^{G}_t, & z_t^{(\ell)}=1,\\
\widetilde{\mathbf{o}}^{L}_t, & z_t^{(\ell)}=0.
\end{cases}
\label{eq:output-fusion}
\end{equation}
Thus, unselected tokens reduce exactly to the local output, while selected tokens form a weighted sum of local and global context. The paired weights explicitly compete over the two spans, unlike additive fusion~\citep{ainslie2023colt5,ren2023sparse}.

\parhead{Budget Control and Progressive Masking}
LoGo controls the global-attention budget by updating the activation threshold from batch-level routing statistics. Given the realized activation ratio $F^{(\ell)}$ and target ratio $\rho^{(\ell)}$ at layer $\ell$, we update:
\begin{equation}
b_{\mathrm{th}}^{(\ell)}
\leftarrow
b_{\mathrm{th}}^{(\ell)}
+\gamma\,\operatorname{sign}\!\left(F^{(\ell)}-\rho^{(\ell)}\right),
\label{eq:threshold-update}
\end{equation}
where $\gamma$ is the threshold update rate. If $F^{(\ell)}>\rho^{(\ell)}$, the threshold increases and fewer tokens activate global attention; otherwise, it decreases. This controller enforces the desired attention budget without auxiliary balancing losses or direct modification of the language-modeling objective~\citep{wang2024auxiliary}. Although layer-specific targets are possible, we use the same target ratio across layers for controlled comparison. 
For example, $\rho^{(\ell)}=0.5$ activates global attention for half of the query tokens in each layer, matching the main global-attention compute of a 1:1 static local-global hybrid.

Because an untrained gate cannot reliably estimate span demand, hard routing from the start may block global-branch learning signals for arbitrary tokens. We therefore use a progressive masking schedule (P-mask): $b_{\mathrm{th}}^{(\ell)}$ is initialized to a negative value so that all tokens initially activate global attention since $p_t^{(\ell)}\in(0,1)$. We set this initial value based on $\gamma$ and the planned P-mask duration, so that $b_{\mathrm{th}}^{(\ell)}$ reaches the initial gate center (0.5) when P-mask ends. As the threshold moves toward the target operating point, global masking is introduced gradually, allowing the gate and both branches to learn from dense supervision before sparse routing takes effect.

\subsection{Efficient Implementation}
\label{sec:efficient-implementation}
LoGo reduces attention FLOPs only when token-level sparsity is exposed to the attention implementation. This section summarizes our parallel kernels, decoding strategy, and complexity analysis, with additional details in Appendix~\ref{app:efficient-implementation}. A naive implementation that computes dense global attention for all tokens and masks outputs would preserve the interface but lose the intended compute benefit. We instead implement the global branch as query-sparse attention: unselected tokens remain in the dense key-value memory but do not instantiate global-query rows, so global-attention cost scales with the number of selected queries.

\parhead{Query-Sparse Prefill and Training}
During prefill and training, the local branch uses standard sliding-window attention. For the global branch, our Triton kernel packs the selected queries into a compacted tensor and computes Eq.~\ref{eq:query-sparse-global}, skipping pruned query rows. The kernel stores the original sequence positions of compacted queries together with compacted \texttt{cu\_seqlens}, which are used to enforce causal visibility and partition KV blocks in the forward pass. The backward pass mirrors the same row-sparse structure, writing query gradients only for selected rows and accumulating KV gradients over causally valid selected-query blocks. This preserves dense-attention semantics for selected rows while avoiding computation for pruned ones.

\parhead{Autoregressive Decoding}
During decoding, since LoGo computes the gate before launching global attention, unselected tokens skip the global call, while selected tokens attend over the full global cache and fuse both branches as in Section~\ref{sec:logo-attention}. For batched decoding, we use a conservative policy that reuses existing dense decode kernels: the global call is skipped only when the microbatch contains no selected rows; otherwise, we launch a dense call and mask inactive outputs. A specialized serving kernel could further compact active decode rows as in prefill.

\parhead{Complexity and Compatibility}
Ignoring projections and MLPs, dense causal attention costs $\mathcal{O}(T^2Hd_h)$ per layer, while the local branch costs $\mathcal{O}(TwHd_h)$. Since LoGo computes global attention only for selected query rows, its global-branch cost $\mathcal{O}(\sum_{t\in\mathcal{I}^{(\ell)}} t\,Hd_h)$ scales as $\mathcal{O}(\rho^{(\ell)}T^2Hd_h)$ under target global ratio $\rho^{(\ell)}$. The total attention cost is therefore $\mathcal{O}(TwHd_h+\rho^{(\ell)}T^2Hd_h)$, matching the dominant compute of a static hybrid with the same global fraction, while allocating global computation dynamically across tokens. Memory-wise, LoGo keeps a dense KV cache with $\mathcal{O}(T)$ memory, as in standard global attention, but can be combined with memory-saving hybrids by replacing their global-attention layers with LoGo. Finally, LoGo uses query-row sparsity shared across heads and preserves the original key-value layout, avoiding head-wise load imbalance and remaining compatible with tensor- and sequence-parallel implementations.

\vspace{-1mm}

\section{Experiments}
\label{sec:experiments}
Our experiments evaluate LoGo along five axes. First, we test whether token-level span allocation preserves Transformer scaling behavior. Second, in a controlled setting, we compare LoGo with the standard Transformer and matched-budget static local-global hybrids. Third, we verify that the query-sparse implementation translates reduced attention FLOPs into operator-level speedups. Fourth, we ablate the main design choices and test robustness across global ratios and hybrid combinations. Finally, we analyze whether LoGo learns meaningful token-level allocation patterns. Training and evaluation details are provided in Appendix~\ref{app:training-setup} and Appendix~\ref{app:evaluation-setup}.

\subsection{Scaling Property}
\label{sec:scaling}
We first test whether LoGo preserves Transformer scaling as a pretrained backbone. We train LoGo with target global ratio $\rho^{(\ell)}=0.5$ and a full-attention Transformer baseline across five model sizes, from 200M to 3.3B parameters. Both families use the same MHA-SwiGLU architecture~\citep{grattafiori2024llama} and identical training hyperparameters, with a fixed tokens-per-parameter ratio following the Chinchilla compute-optimal principle~\citep{hoffmann2022training}. As shown in Table~\ref{tab:scaling}, LoGo preserves Transformer scaling behavior and consistently matches or improves the full-attention baseline across scales. Detailed training setup is provided in Appendix~\ref{app:training-setup}.

\begin{table}[!htbp]
\caption{\textbf{Scaling comparison between LoGo and the Transformer baseline.} We report training loss, WikiText and Lambada perplexity, and average Commonsense Reasoning accuracy.}
\label{tab:scaling}
\centering
\small
\setlength{\tabcolsep}{4pt}
\begin{tabular}{ll|cccc}
\toprule
\textbf{Params.} & \textbf{Model} & \textbf{Train Loss} $\downarrow$ & \textbf{Wiki.} $\downarrow$ & \textbf{LMB.} $\downarrow$ & \textbf{CSR-Avg.} $\uparrow$ \\
\midrule
200M & Transformer & 2.799 & 43.590 & 92.093 & 38.89 \\
     & LoGo & \textbf{2.795} & \textbf{42.990} & \textbf{89.683} & \textbf{40.36} \\
\midrule
470M & Transformer & 2.461 & 28.015 & 30.787 & 44.27 \\
     & LoGo & \textbf{2.449} & \textbf{27.410} & \textbf{27.052} & \textbf{44.30} \\
\midrule
900M & Transformer & 2.294 & 22.037 & 17.717 & 47.63 \\
     & LoGo & \textbf{2.286} & \textbf{21.649} & \textbf{16.431} & \textbf{48.66} \\
\midrule
1.5B & Transformer & 2.162 & 18.241 & 12.025 & \textbf{52.88} \\
     & LoGo & \textbf{2.156} & \textbf{18.052} & \textbf{11.941} & 52.68 \\
\midrule
3.3B & Transformer & 2.000 & 14.911 & 8.216 & 58.11 \\
     & LoGo & \textbf{1.996} & \textbf{14.785} & \textbf{8.009} & \textbf{58.26} \\
\bottomrule
\end{tabular}
\end{table}

\subsection{Comparison with Static Hybrid Paradigms}
\label{sec:static-hybrids}

We next evaluate the performance-compute trade-off by comparing LoGo with standard Transformers and static local-global hybrids under matched parameters and attention FLOPs.

\parhead{Experimental Setup}
We compare four attention paradigms at the 1.5B scale, using the same architecture and training hyperparameters: a standard Transformer, inter-layer and intra-layer local-global hybrids, and LoGo. The hybrid variants use the same local window size $w=128$ and a matched global ratio of 0.5, corresponding to global span in half of the layers, heads, or tokens. We pretrain all models for 100B tokens at 8k context length and then perform two context-extension stages to 32k and 128k, each for 10B tokens. Detailed setup is provided in Appendix~\ref{app:training-setup}.

\parhead{Language Modeling}
After 100B-token pretraining, we evaluate training loss, evaluation perplexity, and commonsense reasoning. As shown in Table~\ref{tab:lm-static-hybrid}, LoGo achieves the best overall results, outperforming both the full-attention Transformer and static hybrid baselines. The same relative ranking holds after 32k context extension (see Appendix~\ref{app:additional-results}, Table~\ref{tab:lm-static-hybrid-32k}).

\begin{table}[!htbp]
\caption{\textbf{Language modeling and commonsense reasoning comparison.} Attention budget denotes the global-attention ratio. Lower is better for loss and perplexity, and higher is better for accuracy.}
\label{tab:lm-static-hybrid}
\centering
\footnotesize
\setlength{\tabcolsep}{3.2pt}
\resizebox{0.9\textwidth}{!}{%
\begin{tabular}{l|c|ccc|ccccccc}
\toprule
\textbf{Model} & \textbf{Attn} & \multicolumn{3}{c|}{\textbf{Loss / PPL} $\downarrow$} & \multicolumn{7}{c}{\textbf{Commonsense Reasoning} $\uparrow$} \\
\cmidrule(lr){3-5}\cmidrule(lr){6-12}
 & \textbf{Budget} & Loss & Wiki. & LMB. & LMB. & PIQA & Hella. & Wino. & ARC-e & ARC-c & \textbf{Avg.} \\
\midrule
Transformer   & 1.0 & 2.119 & 16.12 & 8.36 & 53.97 & 72.14 & 52.63 & 56.35 & 66.84 & 32.25 & 55.70 \\
inter-layer-H & 0.5 & 2.121 & 16.12 & 8.43 & 54.59 & 72.14 & 53.02 & 57.30 & 66.37 & 33.87 & 56.22 \\
intra-layer-H & 0.5 & 2.122 & 17.78 & 8.48 & 54.08 & 72.36 & 52.72 & 57.14 & 66.08 & 32.17 & 55.76 \\
LoGo          & 0.5 & \textbf{2.112} & \textbf{16.04} & \textbf{7.50} & 56.57 & 71.76 & 53.67 & 57.06 & 66.20 & 32.51 & \textbf{56.30} \\
\bottomrule
\end{tabular}}
\end{table}

\parhead{Short-Context Recall}
We further evaluate the 32k checkpoints on real-world recall tasks with relatively short inputs, typically a few thousand tokens. As shown in Table~\ref{tab:short-recall}, LoGo remains competitive with the full-attention Transformer and the intra-layer hybrid, while the inter-layer hybrid shows the only noticeable gap.

\begin{table}[!htbp]
\caption{\textbf{Short-context recall comparison.} We evaluate the 32k checkpoints. Higher is better.}
\label{tab:short-recall}
\centering
\footnotesize
\setlength{\tabcolsep}{4pt}
\begin{tabular*}{0.85\textwidth}{@{\extracolsep{\fill}}l|c|cccccc|c}
\toprule
\textbf{Model} & \textbf{Attn Budget} & \textbf{FDA} & \textbf{SWDE} & \textbf{SQD} & \textbf{TQA} & \textbf{NQ} & \textbf{Drop} & \textbf{Avg.} \\
\midrule
Transformer   & 1.0 & 81.67 & 84.70 & 52.38 & 62.68 & \textbf{33.86} & 29.80 & 57.52 \\
inter-layer-H & 0.5 & 78.49 & 84.34 & 51.41 & 62.32 & 32.34 & \textbf{29.90} & 56.47 \\
intra-layer-H & 0.5 & 83.39 & \textbf{86.32} & \textbf{52.78} & \textbf{63.27} & 32.34 & 28.89 & \textbf{57.83} \\
LoGo          & 0.5 & \textbf{83.85} & 86.05 & 51.94 & 62.32 & 33.26 & 28.46 & \underline{57.65} \\
\bottomrule
\end{tabular*}
\end{table}

\parhead{Long-Range Recall}
To isolate long-range retrieval ability, we evaluate 32k and 128k context-extended checkpoints on needle-style RULER tasks. Table~\ref{tab:long-recall} reports average scores over the subtasks. LoGo achieves the best results in both stages, with larger gains at longer sequence lengths. On RULER-32k and RULER-64k, LoGo obtains the best score on five of the eight subtasks. This trend differs from short-context recall: the inter-layer hybrid becomes comparable to the full-attention Transformer, while the intra-layer hybrid exhibits a clear gap.

\begin{table}[!htbp]
\caption{\textbf{Long-range recall on needle-style RULER tasks.} We evaluate the 32k and 128k checkpoints on eight subtasks, covering three single-needle and five multi-needle variants. Each entry reports the average score for a train-test length setting; per-task results are in Tables~\ref{tab:ruler-32k-subtasks} and~\ref{tab:ruler-128k-subtasks}.}
\label{tab:long-recall}
\centering
\footnotesize
\setlength{\tabcolsep}{10pt}
\begin{tabular}{l|ccc|c|cc|c}
\toprule
\textbf{Model} & \multicolumn{4}{c|}{\textbf{32K CT}} & \multicolumn{3}{c}{\textbf{128K CT}} \\
\cmidrule(lr){2-5}\cmidrule(lr){6-8}
 & \textbf{8k} & \textbf{16k} & \textbf{32k} & \textbf{Avg.} & \textbf{32k} & \textbf{64k} & \textbf{Avg.} \\
\midrule
Transformer   & 89.9 & 81.9 & 64.9 & 78.9 & 73.3 & 42.8 & 58.1 \\
inter-layer-H & 90.2 & 83.9 & 66.5 & 80.2 & 72.1 & 44.9 & 58.5 \\
intra-layer-H & 90.8 & 79.5 & 57.2 & 75.8 & 61.3 & 34.0 & 47.6 \\
LoGo          & \textbf{91.9} & \textbf{85.2} & \textbf{72.0} & \textbf{83.0} & \textbf{78.3} & \textbf{52.5} & \textbf{65.4} \\
\bottomrule
\end{tabular}
\end{table}

Overall, LoGo improves the performance-compute trade-off under matched parameters and attention FLOPs. It maintains competitive language-modeling and short-context recall performance, while providing clear gains on long-range retrieval, supporting the hypothesis that token-level span allocation shifts global computation toward tokens that actually require long-range context.

\subsection{Efficiency}
\label{sec:efficiency}

We benchmark whether LoGo's query-sparse implementation converts reduced attention FLOPs into practical operator speedups, using a Triton kernel built on the FLA baseline~\citep{yang2024fla}. We measure forward-plus-backward runtime from 4k to 64k sequence length, comparing against dense FlashAttention implementations in Triton (the FLA baseline) and CUDA.

\begin{figure}[H]
\centering
\begin{minipage}[t]{0.43\textwidth}
    \centering
    \includegraphics[width=\linewidth]{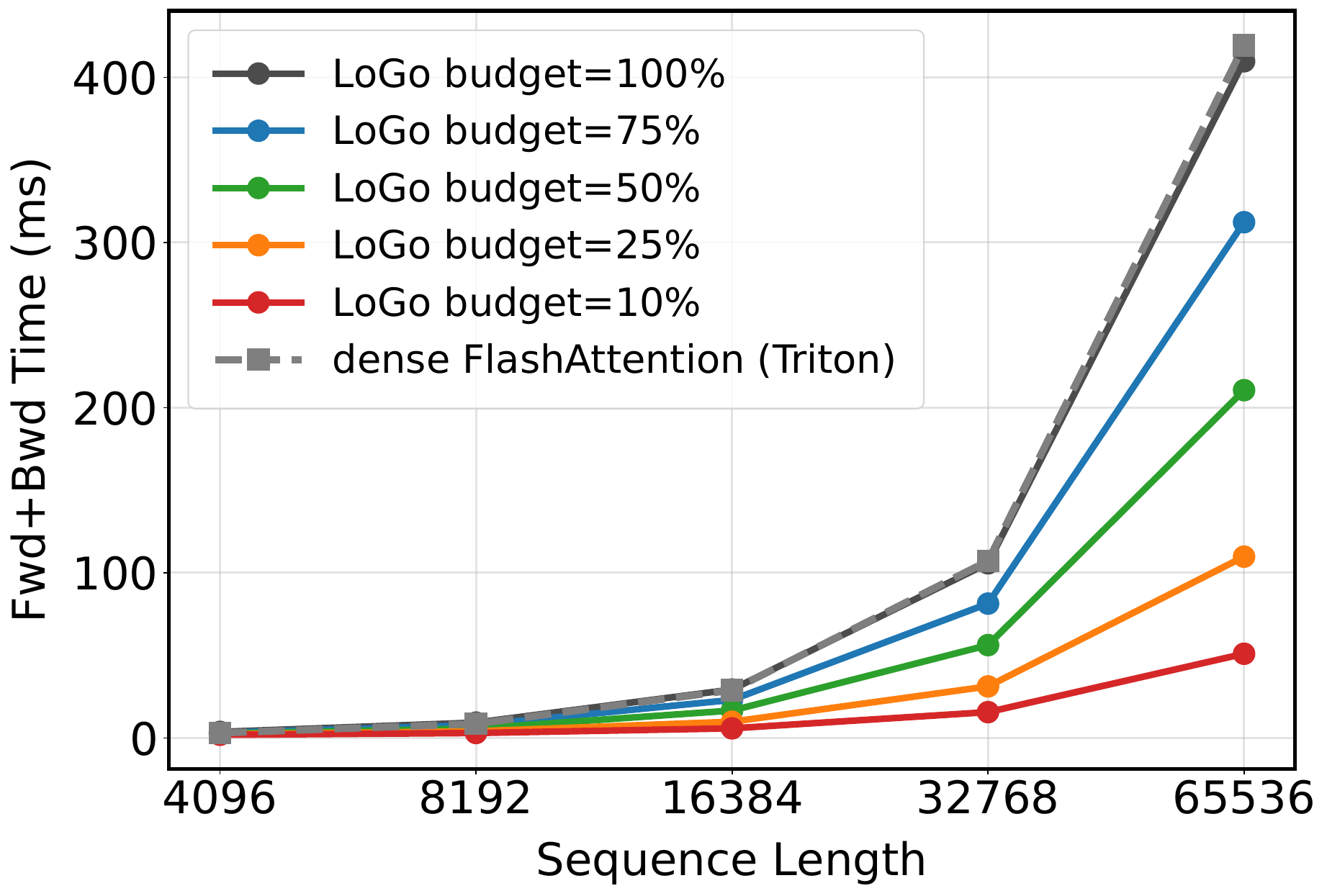}
\end{minipage}
\hspace{0.025\textwidth}
\begin{minipage}[t]{0.43\textwidth}
    \centering
    \includegraphics[width=\linewidth]{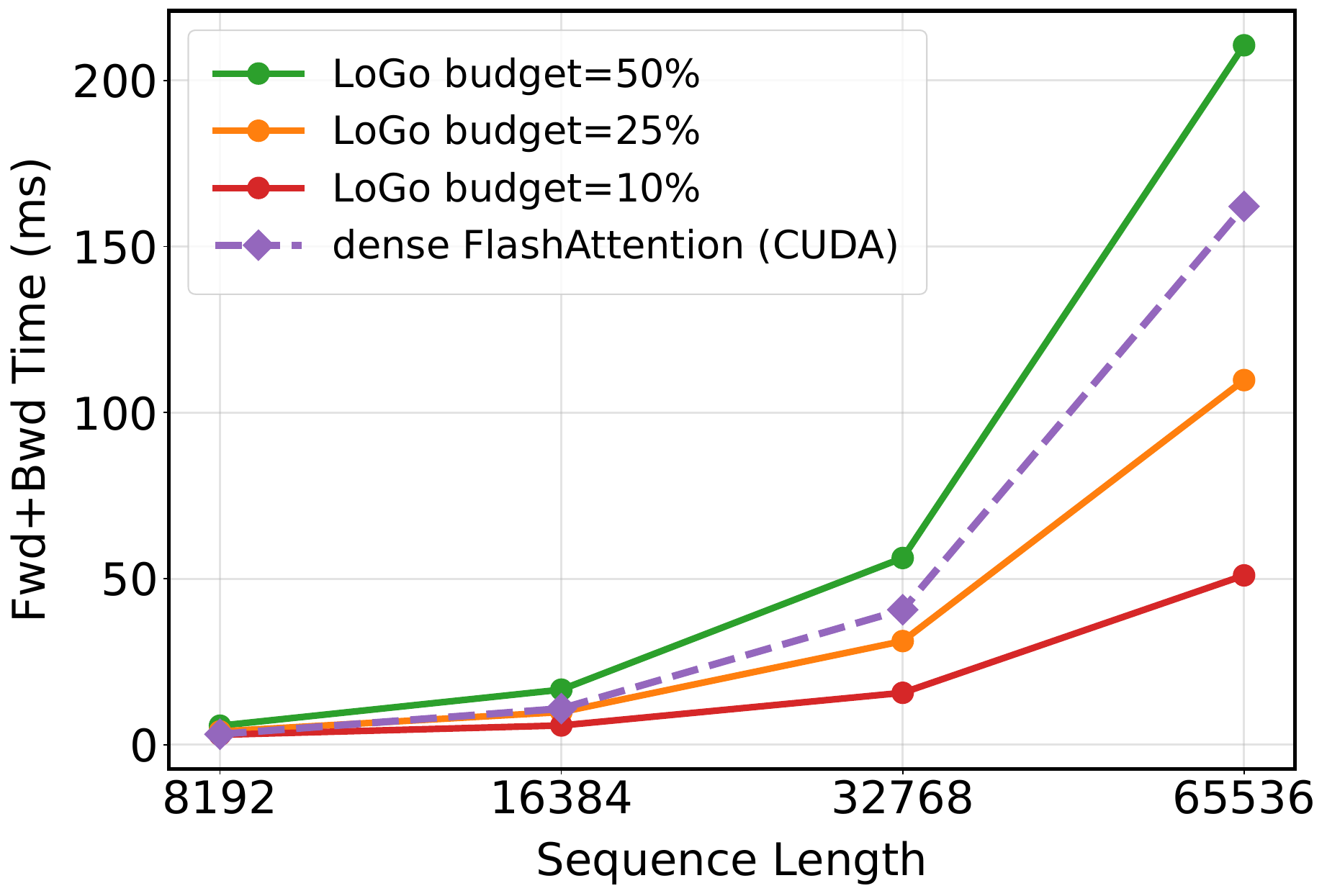}
\end{minipage}
\caption{\textbf{Operator-level efficiency of query-sparse global attention.} Forward-plus-backward runtime is measured with batch size 1, 8 attention heads, and head dimension 128. \emph{Left}: relative to dense FlashAttention in Triton, speedups scale with query sparsity. \emph{Right}: compared against dense FlashAttention in CUDA, our Triton kernel is faster only at high sparsity, leaving room for a CUDA implementation.}
\label{fig:efficiency}
\end{figure}

As shown in Figure~\ref{fig:efficiency}(a), the query-sparse kernel introduces little overhead relative to the dense Triton baseline, and its speedups scale with sparsity across all tested lengths. At a 0.5 attention budget, the measured speedup reaches 1.99$\times$ at 64k, close to the ideal reduction. Figure~\ref{fig:efficiency}(b) compares against the dense CUDA baseline. Due to the implementation gap between Triton and CUDA baselines, the current query-sparse kernel is faster only at high sparsity. Overall, these results show that query-side span sparsity can translate proportional FLOPs savings into practical attention speedups.

\subsection{Ablation Study}
\label{sec:ablation}

\parhead{Architecture Ablations}
We ablate LoGo's main architectural choices using the 1.5B pretraining setting. All variants follow the same training recipe and differ from LoGo by only one design choice. Table~\ref{tab:arch-ablation} reports the results. For branch parameterization, fully sharing branches hurts performance, while fully decoupling them adds substantial parameters and improves perplexity but not recall. LoGo's lightweight decoupling therefore gives the best trade-off. The remaining ablations show that ContextNorm is critical for stable fusion, weighted fusion is preferable to additive fusion, threshold-based budget control outperforms an auxiliary-loss alternative, and P-mask improves downstream performance. Additional interpretation is provided in Appendix~\ref{app:additional-ablations}.

\begin{table}[!htbp]
\caption{\textbf{Architecture ablations.} Rows are grouped by the ablated component. ``--'' indicates a non-converged run.}
\label{tab:arch-ablation}
\centering
\scriptsize
\setlength{\tabcolsep}{3.8pt}
\begin{tabular}{l|l|ccccc}
\toprule
\textbf{Component} & \textbf{Variant} & \textbf{Train Loss} $\downarrow$ & \textbf{Wiki.} $\downarrow$ & \textbf{LMB.} $\downarrow$ & \textbf{CSR-Avg.} $\uparrow$ & \textbf{Recall-Avg.} $\uparrow$ \\
\midrule
\rowcolor{black!6}
-- & \textbf{LoGo} & \underline{2.112} & 16.04 & \textbf{7.50} & 56.30 & \textbf{73.74} \\
\midrule
Branch param. & shared params & 2.116 & 16.12 & 7.83 & 55.99 & 69.92 \\
 & full params & \textbf{2.101} & \textbf{15.85} & \underline{7.71} & \textbf{56.76} & \underline{73.42} \\
 & SO transform & 2.114 & 16.07 & 7.75 & 56.21 & 71.33 \\
\midrule
Normalization & w/o ContextNorm & -- & -- & -- & -- & -- \\
 & SO w/o ContextNorm & 2.121 & 16.36 & 8.39 & 55.43 & 70.90 \\
\midrule
Output fusion & additive fusion & 2.113 & 16.14 & 7.90 & \underline{56.43} & 72.52 \\
\midrule
Budget control & auxiliary loss & 2.118 & 16.18 & 7.87 & 56.18 & 71.73 \\
\midrule
Training schedule & w/o P-mask & \underline{2.112} & \underline{15.96} & 7.73 & 56.08 & 71.59 \\
\bottomrule
\end{tabular}
\end{table}

\parhead{Different Local/Global Ratios}
We also vary the global ratio and compare LoGo with inter-layer hybrids under matched budgets, using the same comparison setting as Section~\ref{sec:static-hybrids}. As shown in Tables~\ref{tab:ratio-recall-ablation} and~\ref{tab:ratio-ruler-ablation}, LoGo consistently improves short-context recall and RULER-32k performance across ratios, showing robustness to the global budget.

\parhead{Compatibility with Other Hybrids}
Finally, we test whether LoGo can complement memory-saving hybrids by replacing the global-attention layers of an inter-layer hybrid with LoGo layers. Table~\ref{tab:hybrid-compat} shows that this combination further reduces compute while improving recall, with details in Appendix~\ref{app:hybrid-compat}. Thus, LoGo can be combined with static hybrids to reduce both computation and memory cost.

\subsection{Analysis}
\label{sec:analysis}

\begin{figure}[H]
\centering
\includegraphics[width=\textwidth]{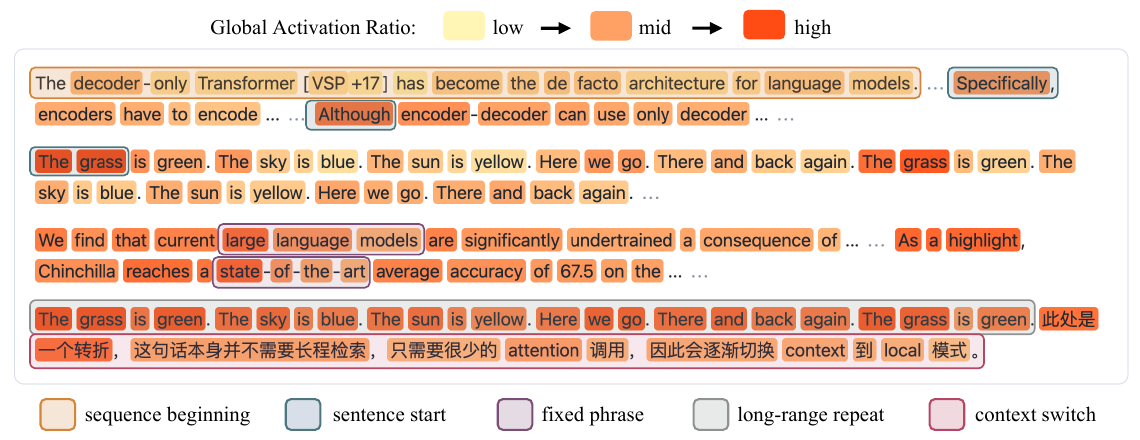}
\caption{\textbf{Word-level span allocation patterns.} Warmer colors indicate more layers activating global attention. Boxed regions mark the five representative patterns in reading order.}
\label{fig:word-level-visualization}
\end{figure}

\parhead{Span Allocation Statistics}
We analyze learned span allocation on evaluation sets from four domains: Wiki (general text), Quest (long-context), GSM8K (math), and the LoGo codebase (code). As shown in Appendix~\ref{app:span-allocation}, the average global ratio stays close to the target 0.5, while allocation remains domain- and token-dependent. Different domains allocate global span at different depths, and tokens within the same domain vary substantially: some use global span in almost no layers or almost all layers, while most fall between these extremes. These patterns indicate that LoGo learns non-uniform span allocation under a controlled global budget.

\parhead{Word-Level Visualization}
We further inspect word-level allocation on a mixed passage with controlled repetitions and a bilingual context switch. Figure~\ref{fig:word-level-visualization} highlights five representative patterns: (1) sequence beginnings mostly remain local; (2) sentence starts often receive more global span; (3) fixed phrases shift toward local attention as later words become predictable from nearby context; (4) long-range repeats trigger strong global activation; and (5) context switches quickly reduce global-span usage. These qualitative patterns indicate that LoGo learns to spend global attention on tokens with long-range demand while avoiding global computation for locally predictable tokens. Additional analysis and the full visualization are provided in Appendix~\ref{app:word-level-visualization}.

\section{Conclusion}
\label{sec:conclusion}

We introduced LoGo, a token-level dynamic local-global attention mechanism for improving the long-context performance-compute trade-off. LoGo treats attention span as a direct proxy for attention budget, giving every token efficient local attention while selectively activating global attention under a controlled budget. Its lightweight branch decoupling, threshold-based budget controller, progressive masking schedule, and query-sparse kernels make this routing stable, budget-controlled, and efficient. Experiments show that LoGo preserves Transformer scaling behavior and improves over full-attention and static hybrid baselines under matched parameters and computation, with clear gains on long-range retrieval. Ablations and analyses further show that the learned span allocation is robust, interpretable, and compatible with static hybrids. These results suggest that learned token-level span allocation is a scalable way to spend global attention where it is most useful.

\subsection*{Acknowledgments}
We thank Yongqi An, Tianyu Fu, Shurong Wang, and Bojian Yin for insightful discussions and constructive feedback.

\clearpage

\bibliographystyle{plainnat}
\bibliography{main}

\clearpage

\beginappendix

\section{Detailed Related Work}
\label{app:related-work}

\subsection{Conditional Computation in Transformers}
\label{app:related-conditional-computation}

\parhead{Token-Adaptive Expert and Depth Budgets}
Conditional computation improves the performance-compute trade-off by assigning different compute budgets to different tokens~\citep{bengio2013estimating,bengio2015conditional}. Existing approaches can be organized by the axis along which computation is made adaptive. Expert-wise methods build on MoE architectures~\citep{shazeer2017outrageously} and dynamically adjust expert computation, for example through top-$p$ expert allocation~\citep{huang2024harder,yang2024xmoe} or null-expert designs~\citep{zeng2024adamoe,jinmoepp}. These methods mainly target feed-forward or expert computation, whereas LoGo targets attention span and the associated long-context cost. Depth-wise methods adapt the number of layers processed by each token, with representative lines including early-exit mechanisms~\citep{elbayaddepth,schuster2022confident} and mixture-of-depths routing~\citep{raposo2024mixture,fu2025think}. They allocate computation along model depth, while LoGo allocates attention span within every layer.

\parhead{Token-Adaptive Attention Budgets}
Attention-wise conditional computation is closest to LoGo. Some methods adapt attention by selecting heads~\citep{zhang2022mixture} or assigning token-wise sparse attention budgets~\citep{lin2026twilight,niu2026stem}. These approaches show that attention computation need not be uniform across tokens. LoGo focuses on attention span because it directly determines both accessible context and attention cost, making it a natural proxy for token-level attention-budget allocation. The most related attention-wise methods dynamically activate specific attention modules. CoLT5~\citep{ainslie2023colt5} routes tokens through light and heavy branches for both feed-forward and attention computation, improving the quality-efficiency trade-off. However, it is designed for an encoder-decoder architecture and applies dynamic computation on the encoder side, whereas LoGo targets decoder-only pretrained backbones and autoregressive long-context modeling. SMA~\citep{ren2023sparse} activates a GAU-style attention module based on SSM states, achieving efficient sequence modeling, but activated tokens still share the same restricted sparse attention pattern. Its router therefore decides whether to enter an attention module rather than which context span a token should receive.

\parhead{Difference from LoGo}
LoGo differs from these conditional-computation methods in three ways. First, it uses attention span as a direct proxy for attention budget, routing each token between local-only and local-plus-global computation. Second, it combines lightweight branch decoupling with an auxiliary-loss-free threshold controller, enabling comparisons under matched parameters and attention budgets. Third, we evaluate this design at pretrained scale with matched parameters and FLOPs, emphasizing long-context retrieval rather than only short-context quality or inference speed.

\subsection{Local-Global Hybrid Attention}
\label{app:related-local-global-hybrids}

\parhead{Inter-Layer Hybrids}
Local-global hybrid attention reduces long-context cost by combining restricted-span computation~\citep{beltagy2020longformer,dai2019transformer,jiang2023mistral7b} with full-context access~\citep{vaswani2017attention}. A common design is the inter-layer hybrid, which assigns different attention mechanisms to different layers. Recent systems alternate global attention with local attention~\citep{team2024gemma,agarwal2025gpt,xiao2026mimo}, linear attention~\citep{li2025minimax,team2025kimi}, or SSM-style modules~\citep{lieber2024jamba,glorioso2024zamba,ren2025samba}. This layer-wise mixture is efficient and simple to scale, but all tokens in the same layer share the same span decision. As a result, tokens that only require local evidence may still pass through global layers, while tokens requiring long-range retrieval cannot receive additional global computation in local layers.

\parhead{Intra-Layer Hybrids}
Another common design is the intra-layer hybrid, which mixes mechanisms within a layer, typically by assigning different heads to different attention types~\citep{donghymba,zuo2025falcon,fu2024mixture}. Head-wise mixtures provide both local and global capacity in each layer and are often considered more capable than layer-wise alternation. However, the span allocation remains static: the same subset of heads is global for every token, regardless of token-level contextual demand. LoGo instead shares each token's routing decision across heads for implementation efficiency, while allowing different tokens to activate global attention in different numbers of layers.

\parhead{Sequence-Adaptive Hybrids}
A smaller set of sequence-adaptive hybrids varies the attention pathway across positions. Some methods use fixed predefined patterns, giving different positions different pathways without learning token-specific budgets~\citep{zhang2025lolcats,mcdermott2025lola,zhang2026sla}. More closely related methods such as TransMamba~\citep{li2026transmamba} and Oryx~\citep{li2026multi} assign different per-token budgets across a sequence, but the allocation is largely handcrafted rather than learned jointly with the language-modeling objective. These designs suggest that sequence-wise variation in attention is useful for efficient context modeling. LoGo instead learns the local/global span decision from token representations, rather than prescribing per-position budgets, allowing global attention to be allocated dynamically under a controlled budget.

\parhead{Difference from LoGo}
LoGo is complementary to static local-global hybrids rather than a replacement for all hybrid designs. It keeps the attention form fixed and learns span allocation end-to-end: every token receives local attention, and a learned gate decides whether to additionally activate global attention under a controlled budget. This makes LoGo directly comparable with inter-layer and intra-layer hybrids under matched global-attention ratios. It also makes LoGo compatible with memory-saving hybrids: global-attention components in a static hybrid can be replaced by LoGo layers to preserve the memory benefits of the underlying layout while reducing global computation.

\section{Efficient Implementation Details}
\label{app:efficient-implementation}

\parhead{Query-Sparse Prefill and Training}
During prefill and training, the global branch uses a selected-query full-causal attention kernel. Selected queries are packed in ascending sequence order into a variable-length compacted layout, while keys and values remain dense over the original sequence. For each compacted query row, the kernel stores its original sequence position, together with compacted query offsets and dense key-value offsets. Causal visibility is therefore determined by real positions rather than compacted indices. In the forward pass, each compacted query block has a known range of original positions. KV blocks before the minimum selected position in the block are fully visible and are processed without a causal predicate, blocks after the maximum selected position are skipped, and only the diagonal region applies a real-position causal mask. This gives the same result as dense full-causal attention on the selected rows while avoiding the significant overhead of applying dense causal masks to every KV block.

The backward pass preserves the same row-sparse structure. The query-gradient kernel mirrors the forward traversal over visible KV blocks and writes gradients only for selected query rows. Gradients for unselected query rows remain zero and are scattered back to the original layout. For key and value gradients, kernels are scheduled over dense KV blocks. A lightweight precomputed search identifies, for each KV block, the first compacted-query block that can causally attend to it. The kernel skips earlier query blocks and processes only the valid region: at most two partially masked diagonal compacted-query blocks, followed by fully causal query blocks that need no causal predicate. As a result, the query-sparse kernels preserve the outputs and gradients of dense global attention restricted to the selected query rows while reducing both forward and backward computation.

\parhead{Autoregressive Decoding}
During autoregressive decoding, each layer maintains the key-value caches required by the local and global branches. The local branch always attends to at most the most recent $w$ cached tokens. Since the gate is computed before global attention is launched, a token routed to local span can skip the global attention call entirely and use the local output. If the token activates global span, it attends over the full global cache and fuses local and global outputs using the same rule as in Section~\ref{sec:logo-attention}.

Our current batched decoding implementation adopts a conservative policy to avoid modifying the existing decode kernel. If no query in a microbatch activates the global branch, the global call is skipped. Otherwise, the standard dense decode kernel is launched for the microbatch, and inactive rows are masked in the output. This policy captures all-local decoding steps without introducing per-step query compaction overhead. A more specialized serving kernel could compact active decode rows in the same spirit as the prefill kernel, further reducing global decoding computation.

\parhead{Complexity and Compatibility}
Ignoring projections and MLPs, dense causal attention in one layer costs $\mathcal{O}(T^2Hd_h)$, and sliding-window attention costs $\mathcal{O}(TwHd_h)$. LoGo adds the local branch for all tokens but evaluates the global branch only on selected query rows, giving total attention cost $\mathcal{O}(TwHd_h+\rho^{(\ell)}T^2Hd_h)$ under target global ratio $\rho^{(\ell)}$. Since $w$ is small in our setting, LoGo has the same dominant attention-compute order as a static local-global hybrid with global fraction $\rho^{(\ell)}$, but it allocates the global computation dynamically at the token level.

The key-value cache remains dense and linear in context length, as in standard global attention. Therefore, LoGo does not by itself provide proportional memory savings. For memory-constrained settings, LoGo can instead be combined with memory-saving local-global designs by replacing their global-attention components with query-sparse LoGo layers, reducing compute while preserving the memory benefits of the underlying hybrid.

LoGo introduces row sparsity rather than head sparsity or key-value sparsity. All heads in a layer share the same token-level routing decision, so each head processes the same number of global query rows and avoids head-wise load imbalance. Meanwhile, the dense key-value layout preserves sequence order and causal structure. Together, these properties make LoGo compatible with tensor parallelism and sequence-parallel implementations. Although token selection is dynamic, the irregularity is converted before the kernel launch into metadata and a compacted query tensor, so the GPU kernel still operates on regular blocks rather than fine-grained branches.

\section{Experimental Details and Additional Results}
\label{app:experiments}

\subsection{Training Setup}
\label{app:training-setup}
\parhead{Scaling Experiments}
The LoGo models and Transformer baselines use the same backbone configuration and training hyperparameters at each scale, trained on an in-house corpus similar to \citet{seed2025seed-oss}. Table~\ref{tab:scaling-model-hparams} summarizes the model hyperparameters, which follow the scaling setup of \citet{sun2024you}. We set the FFN intermediate dimension to $8d/3$. For LoGo, we use target global ratio $\rho^{(\ell)}=0.5$, sliding-window size $w=128$, threshold update rate $\gamma=0.0005$, and initialize $b_{\mathrm{th}}^{(\ell)}$ according to a 20\% P-mask duration. For training, we fix the tokens-per-parameter (TPP) ratio to 20 and set the maximum sequence length to 8192. The batch size is 0.5M tokens for the first four scales and 1M tokens for the 3.3B scale. We use AdamW with weight decay 0.1. The learning rate peaks at the value shown in Table~\ref{tab:scaling-model-hparams} after a 1\% warmup and then cosine-decays to 10\% of the peak.

\begin{table}[!htbp]
\caption{\textbf{Model hyperparameters for scaling experiments.}}
\label{tab:scaling-model-hparams}
\centering
\small
\setlength{\tabcolsep}{12pt}
\begin{tabular}{lcccc}
\toprule
\textbf{Size} & \textbf{Hidden Dim.} & \textbf{\#Layers} & \textbf{\#Heads} & \textbf{LR} \\
\midrule
200M & 768 & 12 & 12 & $6\times10^{-4}$ \\
470M & 1024 & 24 & 16 & $6\times10^{-4}$ \\
900M & 1536 & 24 & 12 & $6\times10^{-4}$ \\
1.5B & 2048 & 24 & 16 & $3\times10^{-4}$ \\
3.3B & 2560 & 32 & 20 & $3\times10^{-4}$ \\
\bottomrule
\end{tabular}
\end{table}

\parhead{Controlled 1.5B Comparison}
For the controlled 1.5B comparison in Section~\ref{sec:static-hybrids}, all models use the same MHA-SwiGLU backbone as in the scaling experiments and differ only in their attention span pathway. We pretrain each model for 100B tokens with a maximum sequence length of 8k and a global batch size of 4M tokens. We use AdamW with weight decay 0.1 and gradient clipping 1.0. The learning rate linearly warms up to $6\times10^{-4}$ over the first 1\% of training and then cosine-decays to $6\times10^{-5}$. After pretraining, we perform two progressive context-extension stages to context lengths of 32k and 128k, each for 10B tokens. The two stages use cosine learning-rate decay to $1\times10^{-5}$ and global batch sizes of 4M and 8M tokens, respectively. For LoGo, we use a threshold update rate of $\gamma=0.0005$ and initialize $b_{\mathrm{th}}^{(\ell)}$ according to a 20\% P-mask duration. Concretely, for a P-mask duration of $S$ steps, we set $b_{\mathrm{th},0}^{(\ell)}=0.5-\gamma S$. For 100B-token pretraining, 20\% duration corresponds to $S=5000$ steps, giving an initial threshold of -2.0.

\subsection{Evaluation Setup}
\label{app:evaluation-setup}
We evaluate language modeling with training loss and perplexity on WikiText (Wiki.; \citep{merity2017pointer}) and Lambada (LMB.; \citep{paperno2016lambada}). Zero-shot commonsense reasoning (CSR) reports average accuracy over Lambada, PIQA~\citep{bisk2020piqa}, HellaSwag (Hella.; \citep{zellers2019hellaswag}), WinoGrande (Wino.; \citep{sakaguchi2020winogrande}), and ARC-Easy / ARC-Challenge (ARC-e / ARC-c; \citep{clark2018think}). Short-context recall includes FDA~\citep{arora2023language}, SWDE~\citep{lockard2019openceres}, SQuAD (SQD; \citep{rajpurkar2018know}), TriviaQA (TQA; \citep{joshi2017triviaqa}), Natural Questions (NQ; \citep{kwiatkowski2019natural}), and DROP~\citep{dua2019drop}, following the evaluation setting of~\citet{arora2024just}. Long-range recall uses all needle-style RULER~\citep{hsieh2024ruler} tasks: three single-needle variants (S-NIAH) and five multi-needle variants (MK-NIAH, MQ-NIAH, and MV-NIAH). All evaluations are run with \texttt{lm-evaluation-harness}~\citep{gao2021framework}.

\subsection{Additional Experimental Results}
\label{app:additional-results}

Table~\ref{tab:lm-static-hybrid-32k} reports language-modeling and commonsense reasoning results after 32k context extension.

Tables~\ref{tab:ruler-32k-subtasks} and~\ref{tab:ruler-128k-subtasks} provide per-task RULER scores for the 32k and 128k checkpoints.

\begin{table}[!htbp]
\caption{\textbf{Language modeling and commonsense reasoning after 32k context extension.} Attention budget denotes the global-attention ratio. Lower is better for loss and perplexity, and higher is better for accuracy.}
\label{tab:lm-static-hybrid-32k}
\centering
\scriptsize
\setlength{\tabcolsep}{4pt}
\resizebox{0.9\textwidth}{!}{%
\begin{tabular}{l|c|ccc|ccccccc}
\toprule
\textbf{Model} & \textbf{Attn} & \multicolumn{3}{c|}{\textbf{Loss / PPL} $\downarrow$} & \multicolumn{7}{c}{\textbf{Commonsense Reasoning} $\uparrow$} \\
\cmidrule(lr){3-5}\cmidrule(lr){6-12}
 & \textbf{Budget} & Loss & Wiki. & LMB. & LMB. & PIQA & Hella. & Wino. & ARC-e & ARC-c & \textbf{Avg.} \\
\midrule
Transformer   & 1.0 & 1.428 & 16.05 & 9.01 & 53.08 & 70.51 & 52.36 & 57.54 & 66.58 & 31.57 & 55.27 \\
inter-layer-H & 0.5 & 1.424 & 15.93 & 8.75 & 54.28 & 71.71 & 52.43 & 57.85 & 65.66 & 33.02 & 55.82 \\
intra-layer-H & 0.5 & 1.425 & 16.05 & 8.82 & 53.50 & 72.09 & 52.69 & 56.67 & 65.70 & 32.42 & 55.51 \\
LoGo          & 0.5 & \textbf{1.417} & \textbf{15.89} & \textbf{7.92} & 55.89 & 72.14 & 53.24 & 57.38 & 66.12 & 32.34 & \textbf{56.18} \\
\bottomrule
\end{tabular}}
\end{table}

\begin{table*}[!htbp]
\caption{\textbf{Per-task RULER results for 32k checkpoints.} We report needle-style RULER scores at 8k, 16k, and 32k input lengths.}
\label{tab:ruler-32k-subtasks}
\centering
\small
\setlength{\tabcolsep}{4pt}
\begin{tabular}{l|ccc|ccc|ccc}
\toprule
\textbf{Model}
& \multicolumn{3}{c|}{\textbf{S-NIAH-1}}
& \multicolumn{3}{c|}{\textbf{S-NIAH-2}}
& \multicolumn{3}{c}{\textbf{S-NIAH-3}} \\
\cmidrule(lr){2-4}\cmidrule(lr){5-7}\cmidrule(lr){8-10}
& \textbf{8k} & \textbf{16k} & \textbf{32k}
& \textbf{8k} & \textbf{16k} & \textbf{32k}
& \textbf{8k} & \textbf{16k} & \textbf{32k} \\
\midrule
Transformer   & \textbf{100.0} & \textbf{100.0} & 94.6 & \textbf{100.0} & \textbf{100.0} & 95.0 & 98.8 & 98.4 & 83.0 \\
inter-layer-H & \textbf{100.0} & \textbf{100.0} & 97.6 & \textbf{100.0} & \textbf{100.0} & 90.6 & 94.6 & 79.2 & 64.0 \\
intra-layer-H & \textbf{100.0} & \textbf{100.0} & 96.6 & \textbf{100.0} & \textbf{100.0} & 83.8 & 99.0 & 81.6 & 59.2 \\
LoGo          & \textbf{100.0} & \textbf{100.0} & \textbf{99.0} & \textbf{100.0} & \textbf{100.0} & \textbf{99.4} & \textbf{100.0} & \textbf{99.6} & \textbf{92.2} \\
\bottomrule
\end{tabular}

\vspace{1.5mm}
\begin{tabular}{l|ccc|ccc|ccc}
\toprule
\textbf{Model}
& \multicolumn{3}{c|}{\textbf{MK-NIAH-1}}
& \multicolumn{3}{c|}{\textbf{MK-NIAH-2}}
& \multicolumn{3}{c}{\textbf{MK-NIAH-3}} \\
\cmidrule(lr){2-4}\cmidrule(lr){5-7}\cmidrule(lr){8-10}
& \textbf{8k} & \textbf{16k} & \textbf{32k}
& \textbf{8k} & \textbf{16k} & \textbf{32k}
& \textbf{8k} & \textbf{16k} & \textbf{32k} \\
\midrule
Transformer   & 86.4 & 78.8 & 62.0 & 98.0 & 97.6 & 73.6 & 60.8 & 55.6 & 27.2 \\
inter-layer-H & 92.2 & 93.2 & 65.2 & 96.6 & 96.8 & 77.4 & \textbf{63.0} & 55.6 & \textbf{53.0} \\
intra-layer-H & \textbf{93.8} & 82.8 & 55.6 & 94.8 & 75.4 & 46.0 & 58.6 & 54.8 & 49.2 \\
LoGo          & 93.4 & \textbf{93.8} & \textbf{70.8} & \textbf{99.4} & \textbf{99.0} & \textbf{91.0} & 61.4 & \textbf{59.2} & 47.4 \\
\bottomrule
\end{tabular}

\vspace{1.5mm}
\begin{tabular}{l|ccc|ccc}
\toprule
\textbf{Model}
& \multicolumn{3}{c|}{\textbf{MQ-NIAH}}
& \multicolumn{3}{c}{\textbf{MV-NIAH}} \\
\cmidrule(lr){2-4}\cmidrule(lr){5-7}
& \textbf{8k} & \textbf{16k} & \textbf{32k}
& \textbf{8k} & \textbf{16k} & \textbf{32k} \\
\midrule
Transformer   & 86.4 & 72.8 & \textbf{52.3} & 89.2 & 51.7 & 31.5 \\
inter-layer-H & 84.0 & 71.9 & 40.2 & 91.3 & \textbf{74.6} & \textbf{44.2} \\
intra-layer-H & \textbf{88.1} & \textbf{73.5} & 38.6 & 91.6 & 67.6 & 28.6 \\
LoGo          & 86.1 & 68.6 & 49.7 & \textbf{94.6} & 61.7 & 26.3 \\
\bottomrule
\end{tabular}
\end{table*}

\begin{table*}[!htbp]
\caption{\textbf{Per-task RULER results for 128k checkpoints.} We report needle-style RULER scores at 32k and 64k input lengths.}
\label{tab:ruler-128k-subtasks}
\centering
\small
\setlength{\tabcolsep}{7pt}
\begin{tabular}{l|cc|cc|cc}
\toprule
\textbf{Model}
& \multicolumn{2}{c|}{\textbf{S-NIAH-1}}
& \multicolumn{2}{c|}{\textbf{S-NIAH-2}}
& \multicolumn{2}{c}{\textbf{S-NIAH-3}} \\
\cmidrule(lr){2-3}\cmidrule(lr){4-5}\cmidrule(lr){6-7}
& \textbf{32k} & \textbf{64k}
& \textbf{32k} & \textbf{64k}
& \textbf{32k} & \textbf{64k} \\
\midrule
Transformer   & 99.8 & 80.0 & \textbf{99.8} & 59.4 & 93.0 & 43.4 \\
inter-layer-H & \textbf{100.0} & \textbf{95.6} & 98.8 & 53.4 & 77.0 & 35.8 \\
intra-layer-H & 99.8 & 70.4 & 95.0 & 41.8 & 70.2 & 27.6 \\
LoGo          & \textbf{100.0} & 83.4 & 99.6 & \textbf{65.0} & \textbf{96.4} & \textbf{57.6} \\
\bottomrule
\end{tabular}

\vspace{1.5mm}
\begin{tabular}{l|cc|cc|cc}
\toprule
\textbf{Model}
& \multicolumn{2}{c|}{\textbf{MK-NIAH-1}}
& \multicolumn{2}{c|}{\textbf{MK-NIAH-2}}
& \multicolumn{2}{c}{\textbf{MK-NIAH-3}} \\
\cmidrule(lr){2-3}\cmidrule(lr){4-5}\cmidrule(lr){6-7}
& \textbf{32k} & \textbf{64k}
& \textbf{32k} & \textbf{64k}
& \textbf{32k} & \textbf{64k} \\
\midrule
Transformer   & 61.6 & 43.2 & 89.8 & 41.0 & 56.8 & 13.0 \\
inter-layer-H & 72.4 & 47.0 & 88.8 & 46.8 & 53.2 & 28.8 \\
intra-layer-H & 58.6 & 34.4 & 53.6 & 25.4 & 51.0 & 20.4 \\
LoGo          & \textbf{80.8} & \textbf{48.6} & \textbf{97.0} & \textbf{68.2} & \textbf{70.2} & \textbf{34.6} \\
\bottomrule
\end{tabular}

\vspace{1.5mm}
\begin{tabular}{l|cc|cc}
\toprule
\textbf{Model}
& \multicolumn{2}{c|}{\textbf{MQ-NIAH}}
& \multicolumn{2}{c}{\textbf{MV-NIAH}} \\
\cmidrule(lr){2-3}\cmidrule(lr){4-5}
& \textbf{32k} & \textbf{64k}
& \textbf{32k} & \textbf{64k} \\
\midrule
Transformer   & 53.8 & \textbf{35.6} & 31.6 & \textbf{27.2} \\
inter-layer-H & 45.3 & 25.0 & \textbf{41.1} & 27.0 \\
intra-layer-H & 37.2 & 25.3 & 25.2 & 26.4 \\
LoGo          & \textbf{53.9} & 35.5 & 28.6 & 26.8 \\
\bottomrule
\end{tabular}
\end{table*}

\FloatBarrier

\subsection{Additional Ablations and Analysis}
\label{app:additional-ablations-analysis}

\subsubsection{Additional Ablations}
\label{app:additional-ablations}

\parhead{Architecture Ablations}
We provide additional interpretation for Table~\ref{tab:arch-ablation}. \textbf{(1) Branch parameterization.} Fully sharing all branch parameters limits specialization and hurts both modeling and recall performance. Fully decoupling branches increases parameters by about 20\% and improves perplexity, but does not improve recall, likely because the sparse global branch becomes harder to train. Replacing the linear-projection (LP) branch transformation with a scale-and-offset transformation (SO) is also slightly worse, supporting LoGo's lightweight LP decoupling design. \textbf{(2) ContextNorm.} Removing ContextNorm prevents proper convergence, as the two branch outputs have substantially different variances. The SO transformation reduces branch discrepancy, but removing ContextNorm under SO still hurts convergence, showing that explicit output normalization remains important for stable fusion. \textbf{(3) Output fusion.} Weighted fusion outperforms additive fusion, supporting explicit competition between local and global spans. \textbf{(4) Budget control.} Replacing threshold-based control with an auxiliary loss degrades modeling performance and yields a higher training loss, justifying the auxiliary-loss-free controller. \textbf{(5) Progressive masking.} Masking the global branch from the beginning hurts downstream performance, confirming the benefit of early dense branch learning before sparse routing takes effect.

\parhead{Different Local/Global Ratios}
Tables~\ref{tab:ratio-recall-ablation} and~\ref{tab:ratio-ruler-ablation} provide the full ratio-sweep results. Parentheses indicate the difference from the matched inter-layer baseline.

\begin{table*}[!htbp]
\caption{\textbf{Short-context recall under different local/global ratios.} We compare LoGo with inter-layer hybrids under matched global-attention budgets using pretrained checkpoints. Because FDA varies noticeably with the global ratio, we additionally report the average excluding FDA.}
\label{tab:ratio-recall-ablation}
\centering
\footnotesize
\setlength{\tabcolsep}{4.5pt}
\resizebox{\textwidth}{!}{%
\begin{tabular}{l|c|rrrrrr|cc}
\toprule
\textbf{Model} & \textbf{Attn Budget} & \textbf{FDA} & \textbf{SWDE} & \textbf{SQD} & \textbf{TQA} & \textbf{NQ} & \textbf{Drop} & \textbf{Avg.} & \textbf{Avg. (w/o FDA)} \\
\midrule
inter-layer-H & 0.25 & 85.48 & 84.07 & 50.17 & 61.55 & 33.01 & 27.55 & 56.97 & 51.27 \\
inter-layer-H & 0.50 & 79.49 & 84.34 & 51.51 & 63.15 & 33.70 & 28.65 & 56.81 & 52.27 \\
inter-layer-H & 0.75 & 84.12 & 84.79 & 50.97 & 63.68 & 33.86 & 28.27 & 57.62 & 52.32 \\
\midrule
LoGo & 0.25 & 85.21 & 83.53 & 50.80 & 61.67 & 35.76 & 28.03 & \textbf{57.50 (+0.53)} & \textbf{51.96 (+0.69)} \\
LoGo & 0.50 & 84.03 & 85.06 & 52.18 & 64.51 & 34.50 & 28.80 & \textbf{58.18 (+1.37)} & \textbf{53.01 (+0.74)} \\
LoGo & 0.75 & 82.58 & 85.60 & 51.51 & 65.23 & 35.51 & 30.81 & \textbf{58.54 (+0.92)} & \textbf{53.73 (+1.41)} \\
\bottomrule
\end{tabular}
}
\end{table*}

\begin{table}[!htbp]
\caption{\textbf{RULER performance under different local/global ratios.} We compare LoGo with inter-layer hybrids under matched global-attention budgets using 32k checkpoints.}
\label{tab:ratio-ruler-ablation}
\centering
\footnotesize
\setlength{\tabcolsep}{8pt}
\begin{tabular}{l|c|cc}
\toprule
\textbf{Model} & \textbf{Attn Budget} & \textbf{RULER 16k} $\uparrow$ & \textbf{RULER 32k} $\uparrow$ \\
\midrule
inter-layer-H & 0.25 & 83.3 & 67.1 \\
inter-layer-H & 0.50 & 83.9 & 66.5 \\
inter-layer-H & 0.75 & 75.1 & 59.9 \\
\midrule
LoGo & 0.25 & 80.9 (-2.4) & \textbf{73.2 (+6.1)} \\
LoGo & 0.50 & \textbf{85.2 (+1.3)} & \textbf{72.0 (+5.5)} \\
LoGo & 0.75 & \textbf{80.1 (+5.0)} & \textbf{67.0 (+7.1)} \\
\bottomrule
\end{tabular}
\end{table}

\phantomsection
\parhead{Compatibility with Other Hybrids}
\label{app:hybrid-compat}
LoGo is complementary to memory-saving hybrid designs. Starting from a 1:1 inter-layer hybrid (global ratio 0.5), we replace its global-attention layers with LoGo layers and report recall performance and three representative multi-needle tasks in Table~\ref{tab:hybrid-compat}. Adding LoGo further reduces compute while improving average recall (65.6/66.9 vs. 65.1). The 0.5-threshold variant also outperforms an equal-FLOPs inter-layer hybrid with a 1:3 global/local ratio (65.6 vs. 64.7). These results show that LoGo can be combined with memory-saving static hybrids to improve the compute-memory-performance trade-off.

\begin{table*}[!htbp]
\caption{\textbf{Compatibility with memory-saving hybrids.} We evaluate pretrained checkpoints and report recall performance and three representative multi-needle tasks; tasks with smaller differences, such as S-NIAH, are omitted.}
\label{tab:hybrid-compat}
\centering
\scriptsize
\setlength{\tabcolsep}{3pt}
\resizebox{\textwidth}{!}{%
\begin{tabular}{l|c|cccccc|ccc|c}
\toprule
\textbf{Model} & \textbf{Attn Budget} & \textbf{FDA} & \textbf{SWDE} & \textbf{SQD} & \textbf{TQA} & \textbf{NQ} & \textbf{Drop} & \textbf{MK1-8k} & \textbf{MQ-8k} & \textbf{MV-8k} & \textbf{Avg.} \\
\midrule
inter-layer-H (1:3) & 0.25 & \textbf{85.48} & 84.07 & 50.17 & 61.55 & 33.01 & 27.55 & \underline{84.4} & 73.8 & 82.4 & 64.7 \\
inter-layer-H (1:1) & 0.50 & 79.49 & \underline{84.34} & 51.51 & 63.15 & 33.70 & \underline{28.65} & \textbf{85.6} & 77.0 & 82.0 & 65.1 \\
\midrule
inter-layer-H (1:1) + LoGo (0.5) & 0.25 & 79.77 & \textbf{84.79} & \underline{52.28} & \textbf{64.93} & \underline{34.62} & 27.12 & 84.2 & \underline{80.2} & \underline{82.4} & \underline{65.6} \\
inter-layer-H (1:1) + LoGo (0.7) & 0.35 & \underline{82.85} & 84.07 & \textbf{53.25} & \underline{64.75} & \textbf{35.86} & \textbf{28.89} & 83.8 & \textbf{84.2} & \textbf{84.2} & \textbf{66.9} \\
\bottomrule
\end{tabular}}
\end{table*}

\subsubsection{Additional Analysis}
\phantomsection
\label{app:span-allocation}

\parhead{Span Allocation Statistics}
Figure~\ref{fig:span-allocation-stats} provides the full layer-wise and token-wise routing statistics across four domains. The average global ratio stays close to the target ratio of 0.5, with mild variation across domains, confirming that the budget controller maintains the intended operating point. However, the realized allocation is not uniform across layers. Wiki and Quest concentrate more global span near the final layer, GSM8K peaks in the first layer, and code uses the least global span in the final layer, suggesting that different domains use long-range context at different stages of representation processing. Token-level span usage also varies substantially within each domain: some tokens use global span in almost no layers or almost all layers, while most fall between these extremes. This confirms that LoGo learns token-level dynamic allocation rather than merely matching the average budget.

\begin{center}
\refstepcounter{figure}\label{fig:span-allocation-stats}
\small
\setlength{\tabcolsep}{1pt}
\begin{tabular}{cc}
\multicolumn{2}{c}{\textbf{Wiki}} \\
\textbf{Layer-wise allocation} & \textbf{Token-wise allocation} \\
\includegraphics[width=0.52\textwidth]{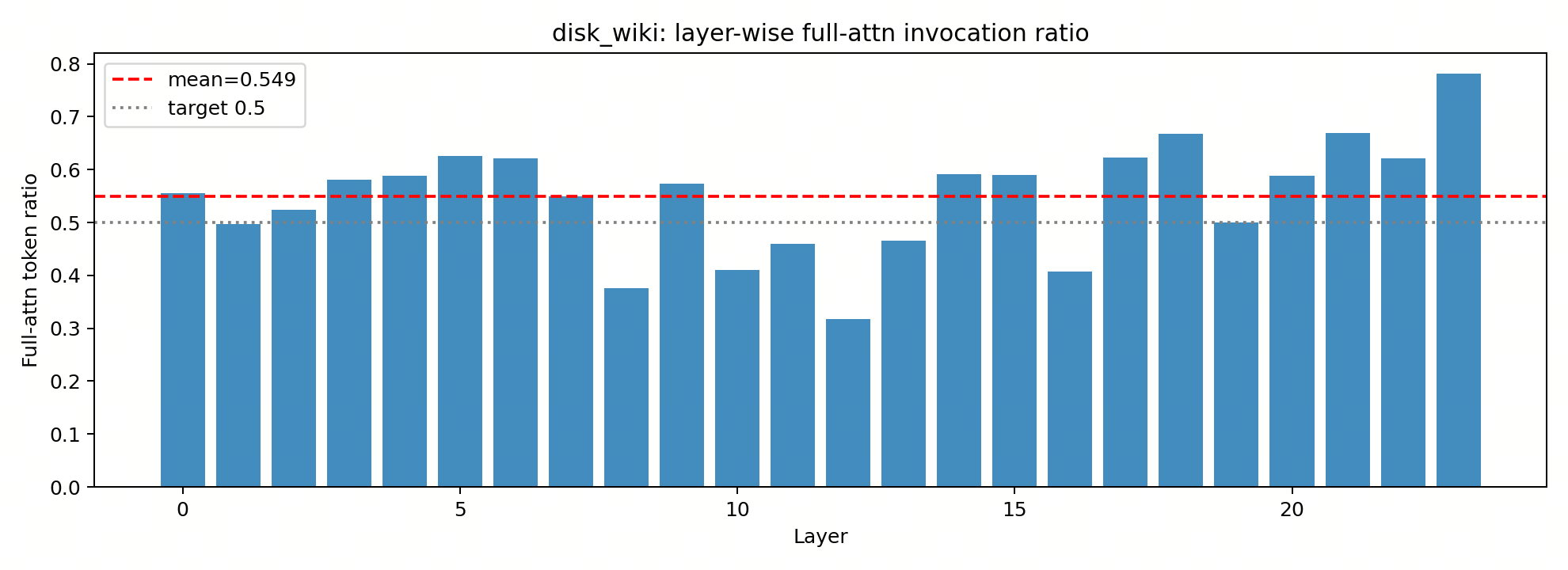} &
\includegraphics[width=0.43\textwidth]{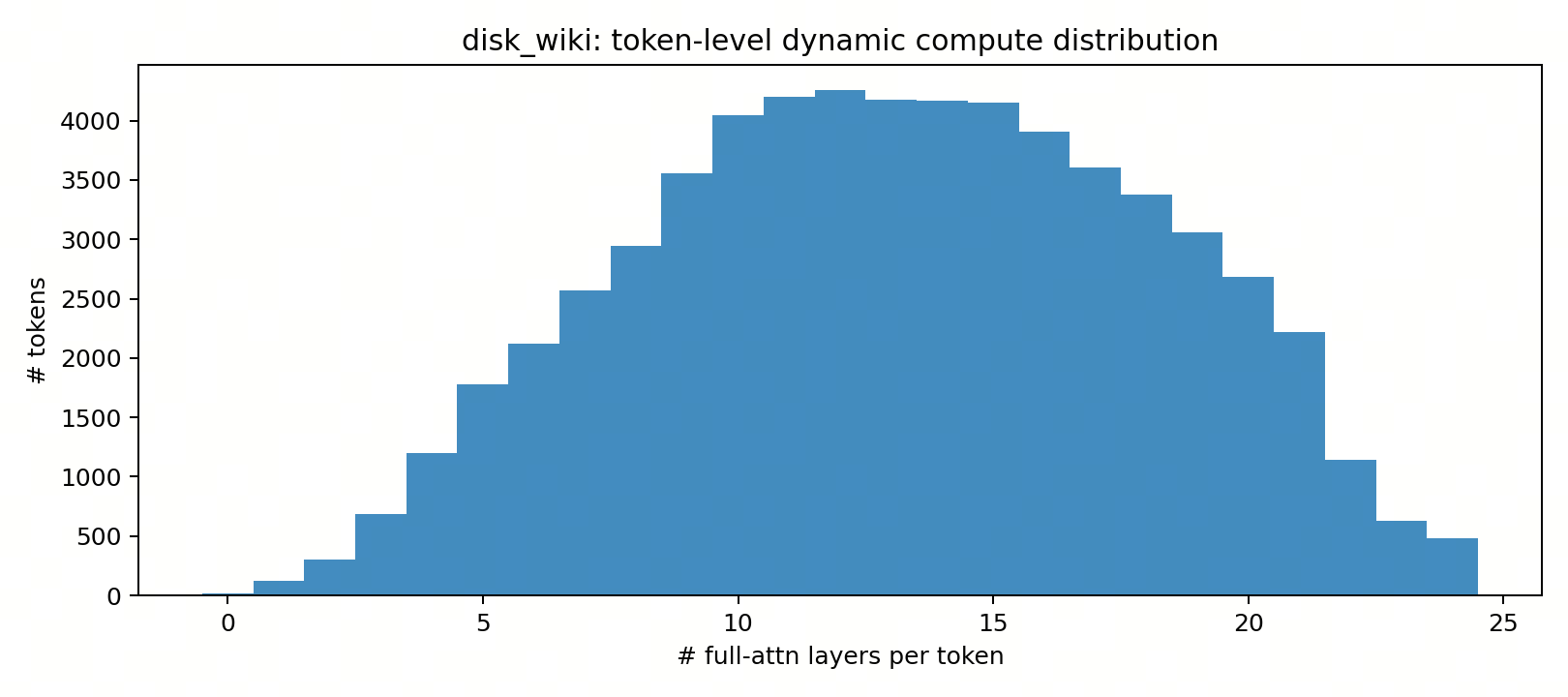} \\
\multicolumn{2}{c}{\textbf{Quest}} \\
\includegraphics[width=0.52\textwidth]{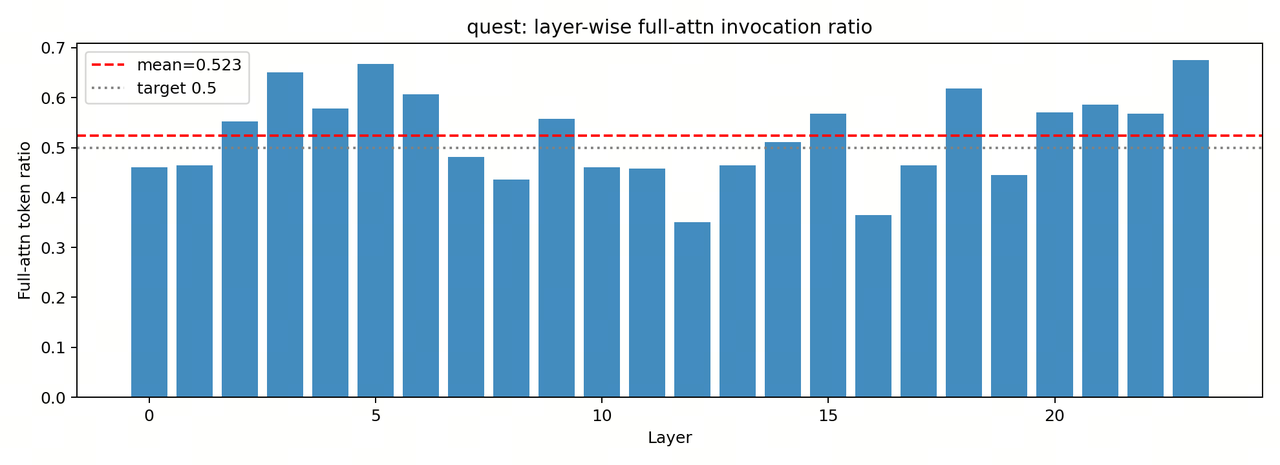} &
\includegraphics[width=0.43\textwidth]{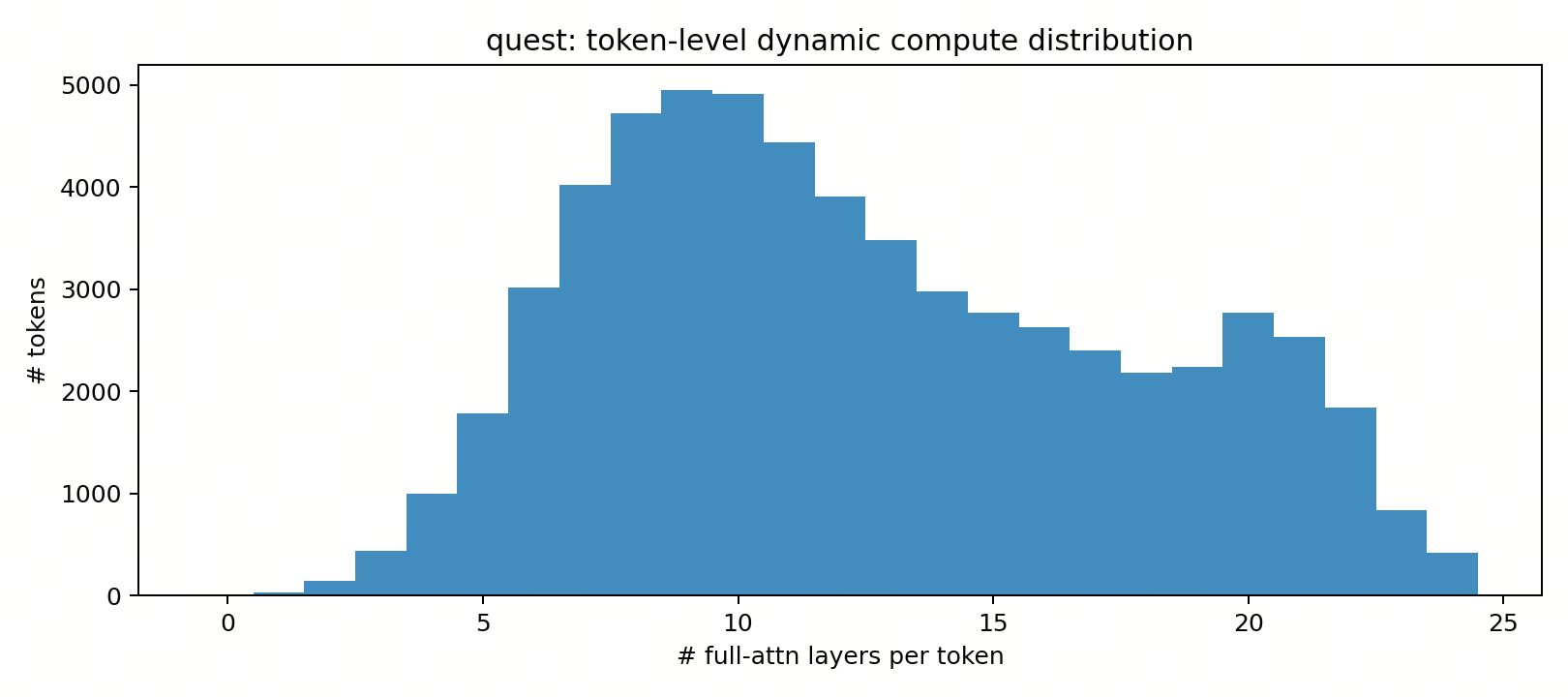} \\
\multicolumn{2}{c}{\textbf{GSM8K}} \\
\includegraphics[width=0.52\textwidth]{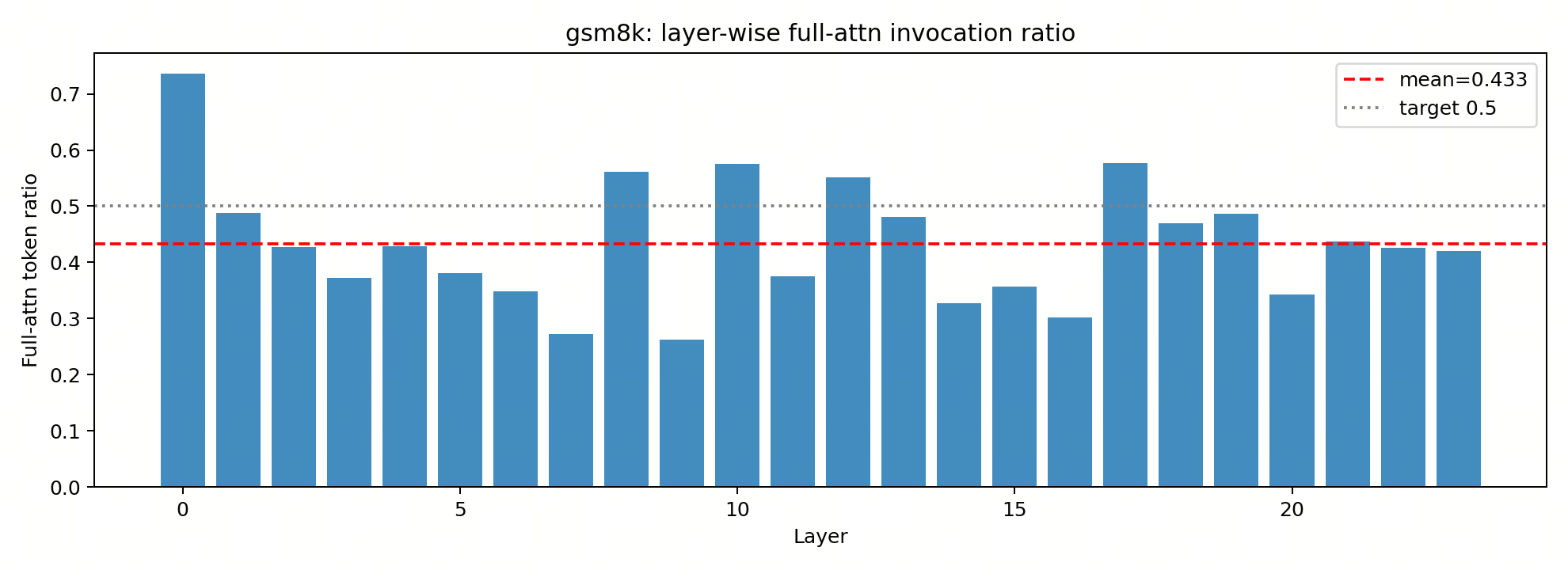} &
\includegraphics[width=0.43\textwidth]{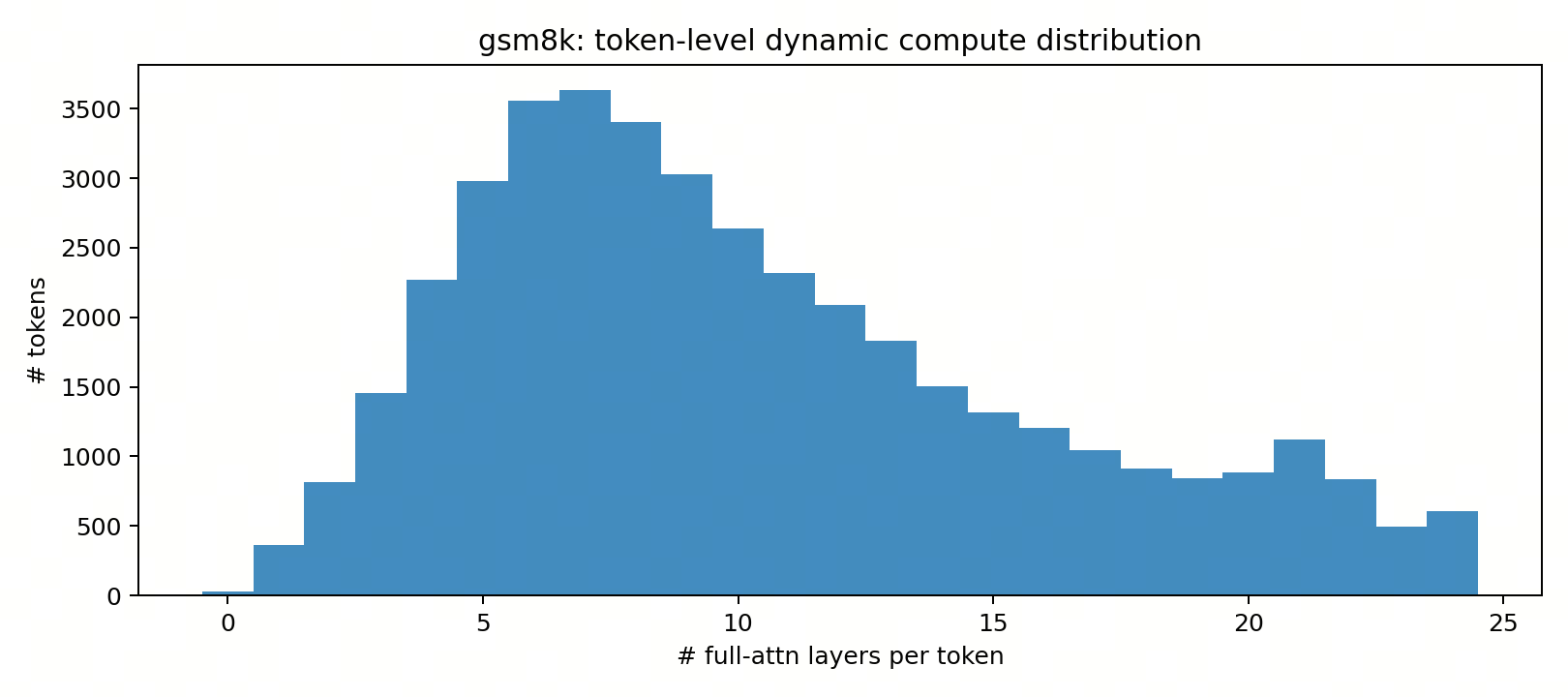} \\
\multicolumn{2}{c}{\textbf{Code}} \\
\includegraphics[width=0.52\textwidth]{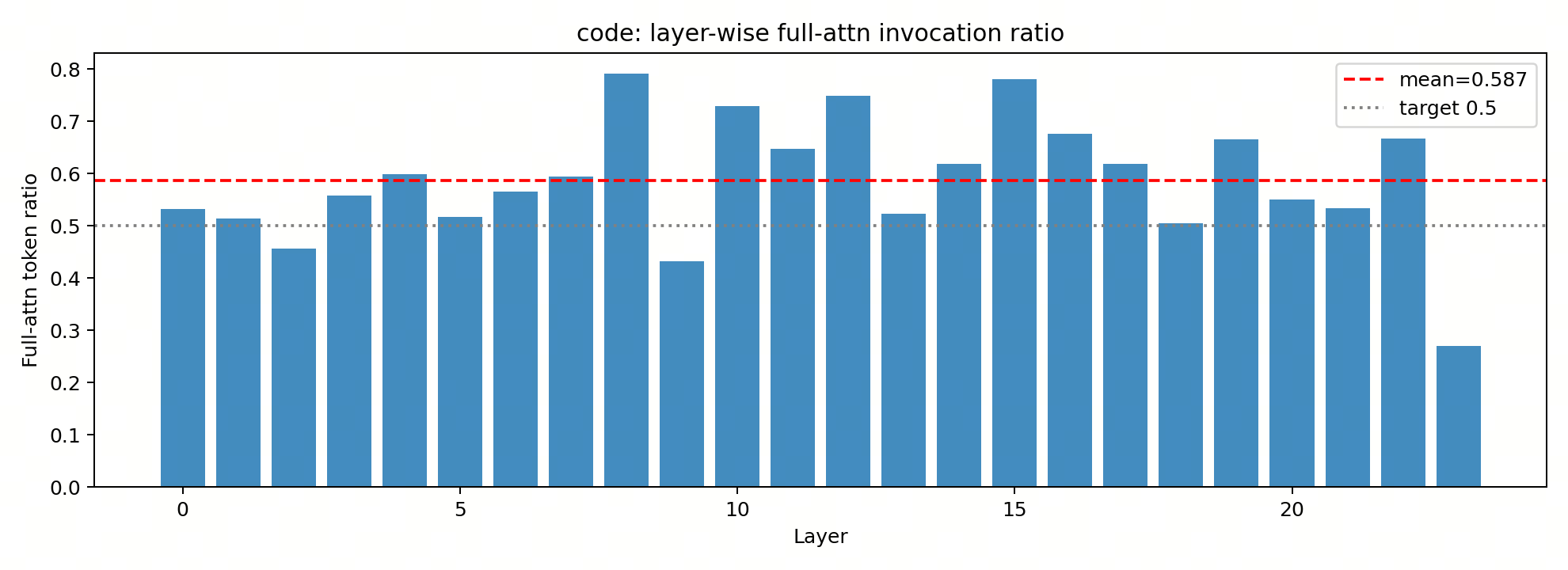} &
\includegraphics[width=0.43\textwidth]{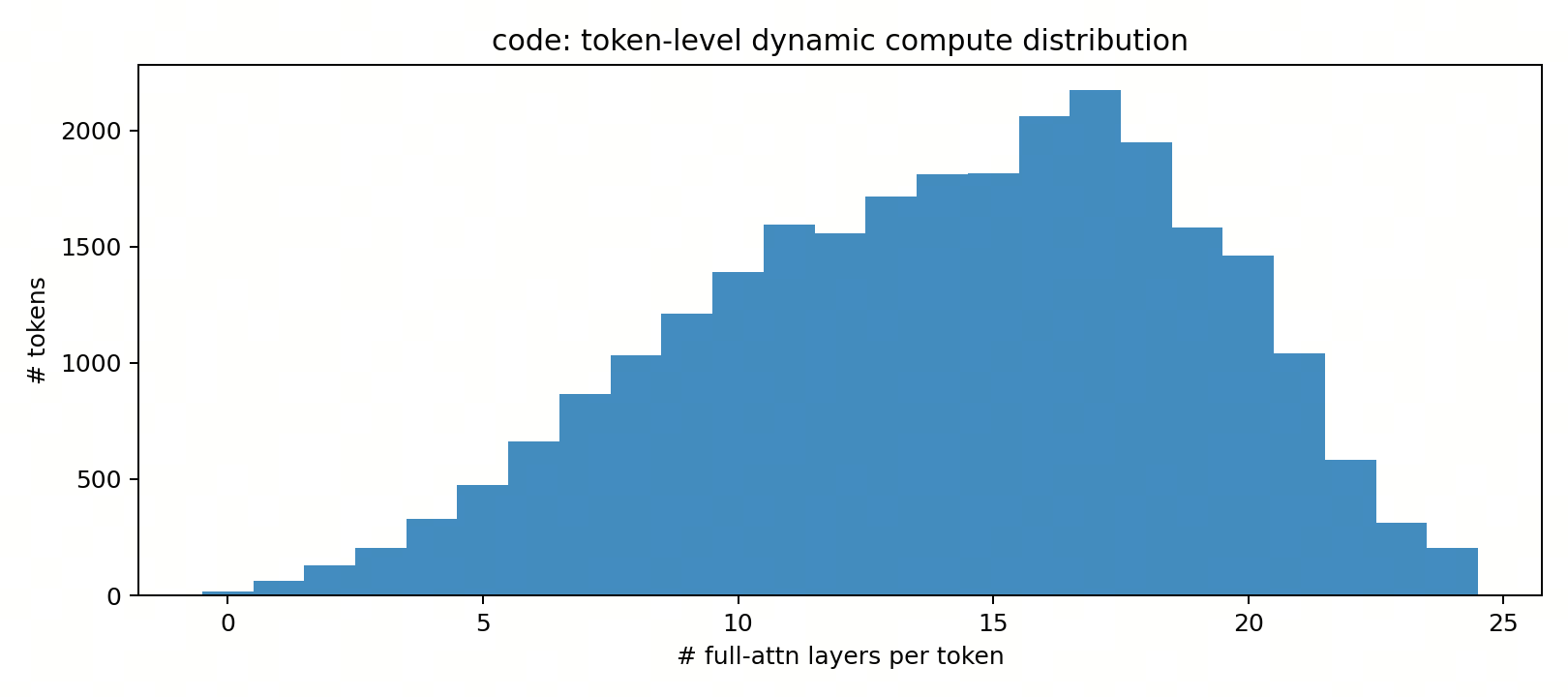} \\
\end{tabular}

\vspace{1mm}
\parbox{0.95\textwidth}{\small \textbf{Figure~\thefigure: Span allocation statistics across domains.} For each domain, the left panel reports the realized global activation ratio by layer, and the right panel reports the distribution of token-level global-span usage, measured by the number of layers with global activation (0--24).}
\end{center}

\clearpage

\begin{figure}[!t]
\centering
\includegraphics[width=\textwidth]{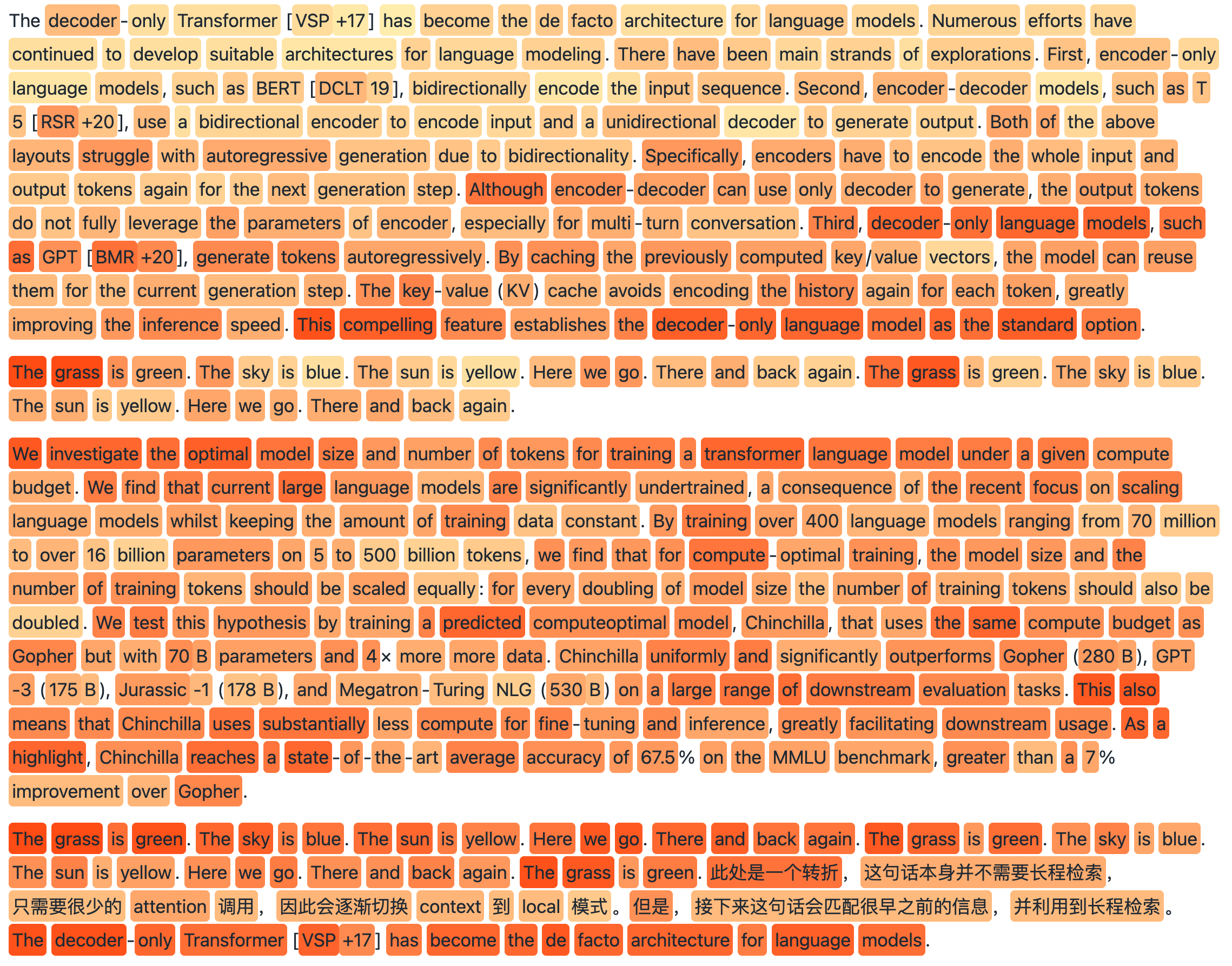}
\caption{\textbf{Full word-level span allocation visualization.} Warmer colors indicate more layers activating global attention.}
\label{fig:word-level-visualization-full}
\end{figure}

\phantomsection
\parhead{Word-Level Visualization}
\label{app:word-level-visualization}
For Figure~\ref{fig:word-level-visualization}, we construct an input passage from YOCO~\citep{sun2024you} and Chinchilla~\citep{hoffmann2022training}, two randomly selected papers cited in this work, and insert two controlled cases: repeated short sentences and a bilingual context switch. Because routing is attached to the query state used to predict the next token, we shift the recorded activation by one position, so each word is colored by the span used to predict it. We then compute word-level global-span activation by averaging over constituent tokens. The full visualization is shown in Figure~\ref{fig:word-level-visualization-full}. The visualization shows five interpretable patterns. \textbf{(1) Sequence beginnings.} Tokens near the beginning of the input sequence mostly use local attention, avoiding unnecessary global-span allocation before long-range context becomes useful. \textbf{(2) Sentence starts.} Sentence-initial tokens after punctuation, especially periods, often receive more global span, consistent with the limited local context available at a new sentence boundary. \textbf{(3) Fixed expressions.} Fixed expressions such as ``large language models'' and ``state-of-the-art'' become increasingly local as later words are locally predictable. \textbf{(4) Long-range repetition.} Repeated content triggers global activation when the match lies outside the local window, whereas nearby repetition uses less global span. We hypothesize that once a few anchor tokens, such as ``The grass is green'', match distant context, subsequent tokens rely on global retrieval to continue the repeated span. \textbf{(5) Context switches.} After switching to a new context, global-span usage drops quickly and rises again only when the passage resumes long-range matching. These patterns provide an interpretable account of how LoGo allocates global computation according to token-level contextual demand.

\end{document}